\documentclass{article}

\usepackage[final, nonatbib]{neurips_data_2023}

\usepackage[utf8]{inputenc} 
\usepackage[T1]{fontenc}    
\usepackage[colorlinks]{hyperref}       
\usepackage{url}            
\usepackage{booktabs}       
\usepackage{amsfonts}       
\usepackage{nicefrac}       
\usepackage{microtype}      
\usepackage{xcolor}         

\usepackage{pifont}
\usepackage{graphicx}
\usepackage{amsmath}
\usepackage{amssymb}

\usepackage{threeparttable}

\usepackage[numbers,sort]{natbib}

\usepackage{wrapfig}
\usepackage{multirow}

\title{SAMRS: Scaling-up Remote Sensing Segmentation Dataset with Segment Anything Model}

\author{%
  Di Wang$^1$\thanks{This work was partially done during Di Wang's internship at iFlytek.},
  Jing Zhang$^2$\thanks{Corresponding author.}~,
  Bo Du$^1$\footnotemark[2],
  Minqiang Xu$^3$,
  Lin Liu$^3$,
  Dacheng Tao$^2$,
  Liangpei Zhang$^4$\footnotemark[2]
  \\
  \textsuperscript{1}School of Computer Science, National Engineering Research Center for Multimedia Software, \\ Institute of Artificial Intelligence, and Hubei Key Laboratory of Multimedia and Network \\ Communication Engineering, Wuhan University, China \\
  \textsuperscript{2}School of Computer Science, Faculty of Engineering, The University of Sydney, Australia \\
  \textsuperscript{3}National Engineering Research Center of Speech and Language Information Processing, China \\
  \textsuperscript{4}State Key Laboratory of Information Engineering in Surveying, Mapping and Remote Sensing, \\ Wuhan University, China
  \\
  \texttt{\{d\_wang,dubo,zlp62\}@whu.edu.cn; jing.zhang1@sydney.edu.au;}\\ \texttt{\{mqxu7,linliu\}@iflytek.com; dacheng.tao@gmail.com}
}

\begin{document}

\maketitle

\begin{abstract}
  The success of the Segment Anything Model (SAM) demonstrates the significance of data-centric machine learning. However, due to the difficulties and high costs associated with annotating Remote Sensing (RS) images, a large amount of valuable RS data remains unlabeled, particularly at the pixel level. In this study, we leverage SAM and existing RS object detection datasets to develop an efficient pipeline for generating a large-scale RS segmentation dataset, dubbed SAMRS. SAMRS totally possesses 105,090 images and 1,668,241 instances, surpassing existing high-resolution RS segmentation datasets in size by several orders of magnitude. It provides object category, location, and instance information that can be used for semantic segmentation, instance segmentation, and object detection, either individually or in combination. We also provide a comprehensive analysis of SAMRS from various aspects.  Moreover, preliminary experiments highlight the importance of conducting segmentation pre-training with SAMRS to address task discrepancies and alleviate the limitations posed by limited training data during fine-tuning. The code and dataset will be available at \href{https://github.com/ViTAE-Transformer/SAMRS}{SAMRS}.
\end{abstract}

\section{Introduction}
\label{sec:intro}

The advancement of earth observation technologies has led to the generation of abundant remote sensing images (RSI). These images retain valuable information about the spatial distribution and condition of extensive ground surfaces and geospatial objects, and can be conveniently accessed in real-time. Consequently, remote sensing data has garnered the interest of various disciplines, including agricultural monitoring, urban planning, and environmental protection. In particular, the identification of surface targets has been a fundamental task in these fields for several years.

To our knowledge, a significant number of RSIs remain unlabeled. Unlike natural images that can be easily comprehended by the human eye, interpreting RSI taken from an aerial perspective typically demands specialized expertise from practitioners. Furthermore, RSI objects are often distributed sparsely, and the images frequently contain small targets, making the labeling process less efficient. Therefore, the annotation of RSI has traditionally required substantial labor and time costs. Among various RS tasks, the classification task requires only a single category for the entire scene, and the detection task involves the additional step of bounding box annotation, while segmentation is particularly challenging since it necessitates pixel-level annotations to accurately delineate object boundaries.

\begin{figure}[t]
    \centering
    \includegraphics[width=\linewidth]{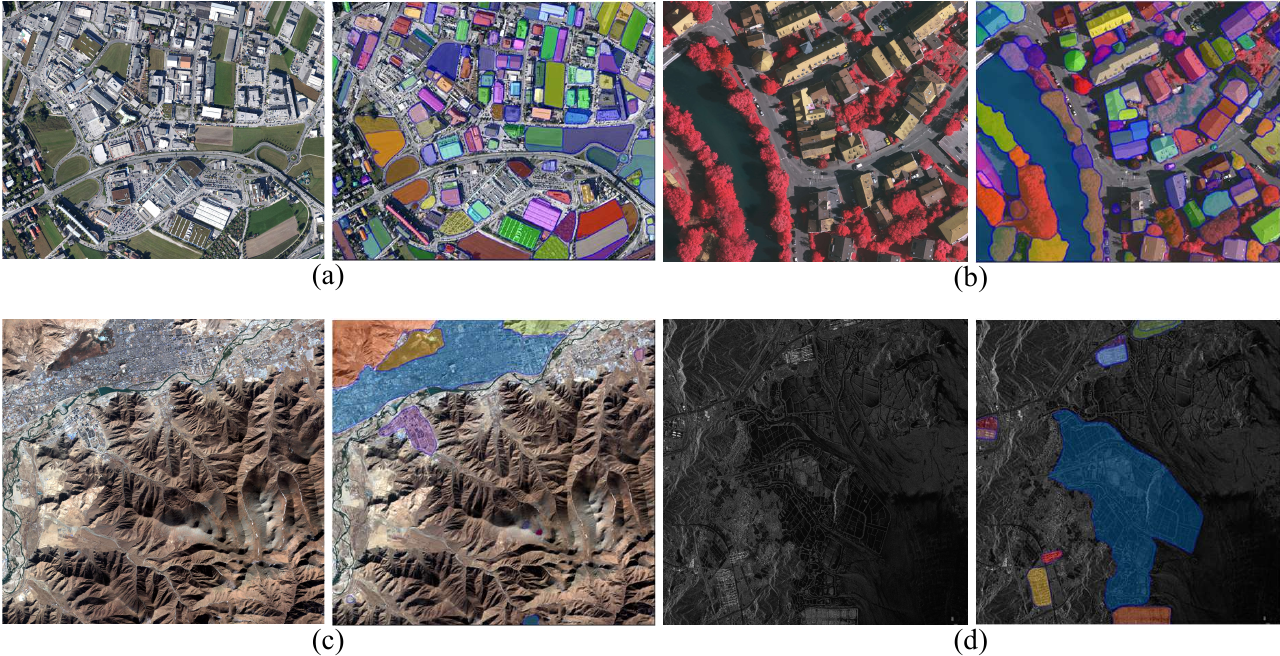}
    \caption{Some examples of SAM segmentation results on RSIs: (a) RGB aerial image obtained from the IsAID dataset \cite{isaid}. (b) Airborne aerial image composed of near-infrared, red, and green bands. This image is from the ISPRS Vaihingen dataset\protect\footnotemark[1]. (c) RGB satellite image observed by GF-2 sensors. This image is from the GID dataset \cite{gid}. (d) Hisea-1 SAR image from the Marine Farms Segmentation track of the 5th Gaofen Challenge\protect\footnotemark[2]. These segmentation results are generated by the SAM demo website\protect\footnotemark[3].}
    \label{sam_example}
\end{figure}
\footnotetext[1]{\url{https://www.isprs.org/education/benchmarks/UrbanSemLab/2d-sem-label-vaihingen.aspx}}
\footnotetext[2]{\url{https://www.gaofen-challenge.com/challenge}}
\footnotetext[3]{\url{https://segment-anything.com/demo}}

Do we have to spend a significant amount of time annotating RSIs? The answer is probably no. Recently, the segment anything model (SAM) \cite{sam}, which excels in object segmentation, has gained popularity as a new research focus in the field of computer vision. SAM accurately captures object locations and contours (\textit{i.e.}, in the form of masks), enabling it to distinguish various objects in the foreground and background. Furthermore, SAM possesses an impressive zero-shot segmentation ability, exhibiting high performance even when applied to specialized scenarios such as cell images photographed by microscopes \cite{cell} and medical images \cite{medical}, despite being trained on a vast dataset of natural images. In the RS field, \cite{sam_space} firstly tests the performance of SAM on six public datasets. \cite{sam_mars} extra introduce a domain decoder to improve the performance of SAM on the planetary geological mapping task. Beyond default prompts, \cite{sam_oneshot,sam_text2seg} consider utilizing texts as the prompt by adopting Grounding DINO \cite{groundingdino} to obtain boxes that can be employed by SAM. Then, \cite{sam_oneshot} realizes the one-shot segmentation with the help of PerSAM \cite{persam}, while \cite{sam_text2seg} applies the heatmap obtained from CLIP \cite{clip} to further optimize segmentation results. Different from the above methods with manual prompts, \cite{rsprompter} design a prompter to adaptively generate prompts for improving the performance of SAM in instance segmentation. In addition, SAM is also used in producing rotated bounding boxes \footnote[4]{\url{https://github.com/Li-Qingyun/sam-mmrotate}}, which is significant for RS oriented object detection.

We have also found it performs well in recognizing diverse targets in RSI, even when the images are obtained using sensors that perceive different bands, such as infrared and microwave, or with varying resolutions, such as airborne or satellite imagery, as illustrated in Figure \ref{sam_example}. Although we acknowledge that SAM may not have fully detected all regions, we believe that it has significant potential to improve the efficiency of annotating RSIs since it delivers promising segmentations on recognized areas. Therefore, in this study, we aim to utilize SAM to efficiently construct a large-scale RS segmentation dataset by obtaining pixel-level annotations for RSIs. Ground objects in RSI possess definite category properties, which are essential for real RS recognition tasks. However, the segmentation maps produced by SAM lack such information, rendering them unsuitable for labeling RSIs. To address this issue, we notice the annotations in existing RS object detection datasets, which include category and bounding box information. With the aid of SAM, we can leverage such detection annotations to obtain pixel-level semantic labels and efficiently construct large-scale segmentation datasets. The obtained dataset is called \textbf{S}egment \textbf{A}nything \textbf{M}odel annotated \textbf{R}emote Sensing \textbf{S}egmentation dataset (SAMRS). SAMRS inherits the characteristics of existing RS object detection datasets that have more samples and categories compared with existing high-resolution RS segmentation datasets.

Since we efficiently obtain numerous segmentation label maps, it is natural to consider using the obtained dataset for pre-training. Existing models pretrained by classification tasks may be not very suitable for downstream tasks, e.g., segmentation, because of the task-level discrepancy \cite{rsp}, while the emergence of SAMRS is expected to address this issue. To this end, we train classical deep learning models on the SAMRS, and finetune the trained model on typical RS segmentation datasets to explore the feasibility of segmentation pre-training. The main contribution of this study can be summarized to: \textbf{(1)} We develop a SAM-based pipeline for efficiently generating RS segmentation annotations. \textbf{(2)} We obtain a large-scale RS segmentation dataset named SAMRS using existing RS object detection annotations, whose capacity is far beyond existing high-resolution RS segmentation datasets. \textbf{(3)} We conduct preliminary segmentation pre-training experiments on SAMRS. The results highlight the importance of conducting segmentation pre-training using large-scale RS segmentation data, such as SAMRS, for mitigating task discrepancy and dealing with limited training data. We hope this research could significantly enhance the annotation efficiency of RSIs, thereby unlocking the full potential of RS models, especially in the context of segmentation tasks. 

\section{Implementation}
\subsection{Segment Anything Model} 

\begin{wrapfigure}[]{r}[0em]{0.5\textwidth}
    \vspace{-10mm}
    \centering
    \includegraphics[width=\linewidth]{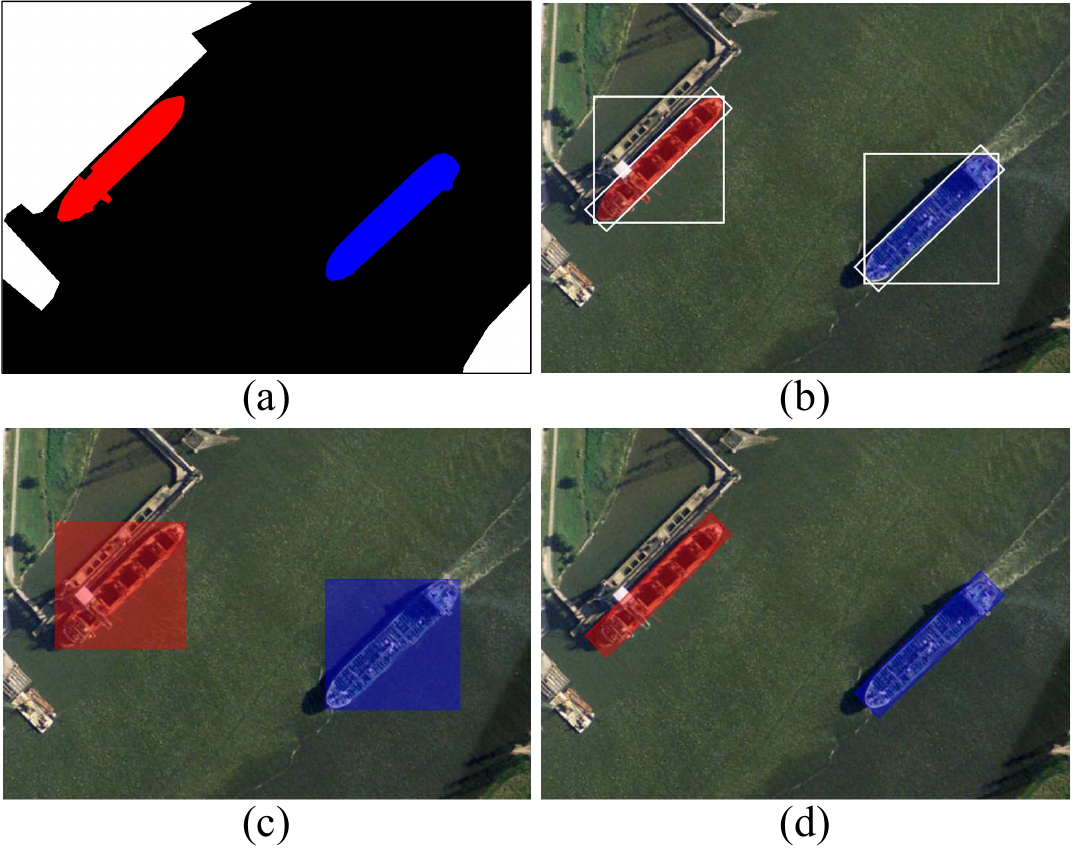}
    \caption{The differences between segmentation labels and mask prompts. (a) Pixel-level annotated map from the original dataset. (b) Pixel-level annotations along with horizontal and rotated box ground truths. (c) Mask prompts derived from horizontal boxes. (d) Mask prompts derived from rotated boxes. The ship instances are marked with different colors by following (a).
    }
    \vspace{-8mm}
    \label{ground_truth}
\end{wrapfigure}

To perform segmentation, additional prompts are needed to guide SAM to locate the object of interest, in addition to the input image. SAM supports various prompts, such as points, boxes, and masks, which can be input into the model either alone or in combination. It is important to note that when using point prompts, it is necessary to indicate whether the points are foregrounds or backgrounds. In this study, we use detection annotations from existing datasets to obtain all kinds of prompts since they contain both location and category information.

\subsection{Datasets}

In this study, we employ SAM on four public RS object detection datasets, namely HRSC2016 \cite{hrsc2016}, DOTA-V2.0 \cite{dotav2}, DIOR \cite{dior}, and FAIR1M-2.0 \cite{fair1m}. DOTA, DIOR, and FAIR1M are three large-scale datasets \cite{fair1m}. HRSC2016 is primarily designed for ship detection and comprises only one category. In comparison to the other three datasets, it has the smallest data volume. Additionally, in the testing set, 124 images possess bounding box annotations and pixel-level labels simultaneously, making it highly suitable for evaluating the accuracy of SAM annotations. Therefore, we conduct an ablation study on the testing set consisting of the aforementioned 124 images to determine the optimal configuration for SAM. Following this, we generate segmentation labels for the remaining datasets. To obtain a segmentation dataset with more images or categories, we opt for the latest versions of DOTA and FAIR1M. Based on the available annotations, we only transform the training and validation sets of DOTA-V2.0 and FAIR1M-2.0, while for DIOR, all data has been utilized. Here, according to the licenses, DOTA, DIOR, and FAIR1M can be used for academic purposes.

\subsection{Prompt Settings}

As RSIs are captured from an overhead perspective, the objects in them can have arbitrary orientations, unlike natural image objects that are typically oriented upward due to gravity. Hence, in addition to the usual horizontal bounding boxes (H-Box), we also consider oriented bounding boxes or rotated bounding boxes (R-Box) as box prompts. However, SAM does not directly support R-Box prompts. To address this issue, we use the minimum circumscribed horizontal rectangle of the R-Box, which is denoted as ``RH-Box''. It is also worth noting that the instances in the HRSC2016 testing set contain both H-Box and R-Box ground truth annotations.

\begin{wrapfigure}[]{r}[0em]{0.5\textwidth}
    \centering
    \includegraphics[width=\linewidth]{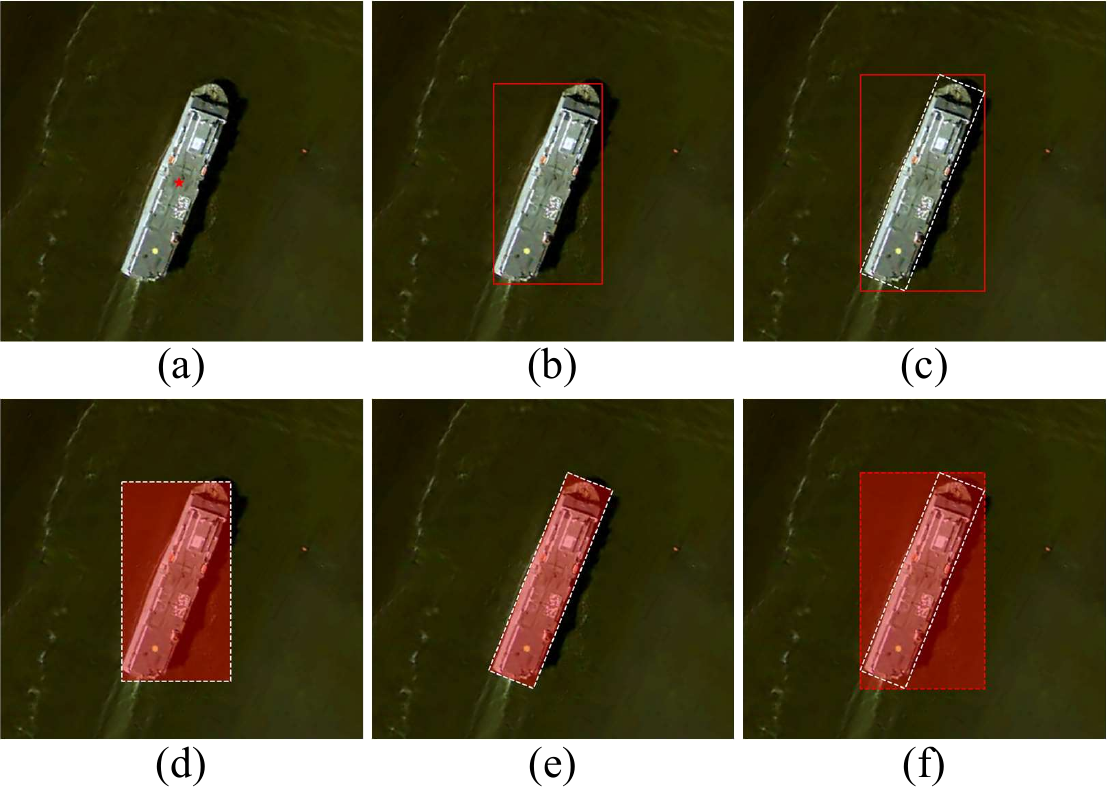}
    \caption{The adopted basic prompts. (a) CP. (b) H-Box. (c) RH-Box. (d) H-Box-M. (e) R-Box-M. (f) RH-Box-M. The dashed line is used for the convenience of visualization.}
    \label{prompt}
\end{wrapfigure}

In the case of the point prompt, due to the intricate shapes of various RS objects, such as airplanes, we have taken a cautious approach and only consider the center point as the foreground. We did not include background points in our study, as accurately defining them in an automated way can be challenging without additional contextual information. Regarding the mask prompt, we define the region enclosed by corresponding boxes as the mask prompt. Figure \ref{ground_truth} illustrates the differences between the adopted mask prompts and ground truth segmentation labels. In SAM, the mask is a single-channel score matrix where positive values denote the active area where the target is located, whereas negative values represent irrelevant areas. In our experiments, we assign the values in these two types of areas as 1,000 and -1,000, respectively.

In summary, we have obtained six basic prompts, namely center point (CP), H-Box, RH-Box, and their corresponding masks, \textit{i.e.}, H-Box-M, R-Box-M, and RH-Box-M, as illustrated in Figure \ref{prompt}.

\subsection{Ablation Study} 
\label{subsec:ablation}

In addition to the above basic prompts, we also investigate various combinations of prompts in this study. To conduct a comprehensive analysis, we compute two types of mean intersection over union (mIOU) metrics: $\text{mIOU}_{I}$ and $\text{mIOU}_{P}$, which measure the similarity between the predicted segmentation mask and the ground truth label. The former is the average value of the IoU calculated on a per-instance basis, while the latter measures the pixel-level accuracy. Given the $i$th instance with intersection set $I_i$ and union set $U_i$, and the number of instances $N$, we have:
\begin{equation}
    \text{mIOU}_{I} = \frac{1}{N} \sum\limits_{i=1}^{N} \frac{I_i}{U_i} \quad \text{mIOU}_{P} = \frac{\sum_{i=1}^{N} I_i}{\sum_{i=1}^{N} U_i}.
\end{equation}

\begin{table}[t]
  \caption{Results of using different prompts on the HRSC2016 testing set consisting of 124 images.}
  \newcommand{\tabincell}[2]{\begin{tabular}{@{}#1@{}}#2\end{tabular}}
  \centering
  \resizebox{0.7\linewidth}{!}{
    \begin{tabular}{cccccc|cc}
  \hline
  CP & H-Box & H-Box-M & R-Box-M & RH-Box & RH-Box-M  & $\text{mIOU}_{I}$ & $\text{mIOU}_{P}$ \\
  \hline
  \bfseries \textit{Point} & \multicolumn{7}{c}{}\\
  \hline
  \ding{51}  &    &    &   &  &    &  16.14 & 2.72 \\
  \hline
  \bfseries \textit{H-Box} & \multicolumn{7}{c}{}\\
  \hline
          &  \ding{51}  &    &   &  &    & \bfseries 89.97 & \bfseries 79.40 \\
          &    &   \ding{51} &   &  &    &  40.54 & 36.71 \\    
 \ding{51} & \ding{51}  &    &   &  &    &  86.67 & 77.35 \\     
         & \ding{51}  &  \ding{51}  &   &  &    &  74.21 & 62.25 \\   
  \ding{51}  &  &  \ding{51}  &   &  &    & 24.54 & 5.41 \\  
  \ding{51}  & \ding{51} &  \ding{51}  &   &  &    & 59.71 & 49.30 \\    
  \hline
  \bfseries \textit{R-Box} & \multicolumn{7}{c}{}\\
  \hline
   &   &    & \ding{51}  &   &    & \bfseries  65.54 & \bfseries  59.78 \\
   \ding{51} &   &    & \ding{51}  &  &    & 26.49 & 4.97 \\   
  \hline
  \bfseries \textit{RH-Box} & \multicolumn{7}{c}{}\\
  \hline
         &   &    &   & \ding{51} &    &  \bfseries 88.85 & \bfseries 76.42 \\  
         &   &    &   &  &  \ding{51}  &  34.63 & 31.81 \\    
  \ding{51}    &   &    &   & \ding{51} &    &  83.55 & 72.67 \\    
   &   &    &   & \ding{51} &  \ding{51}     &  66.23 & 52.75 \\ 
   \ding{51} &   &    &   &  &  \ding{51}     &  23.71 & 5.10 \\ 
   \ding{51} &   &    &   & \ding{51} &  \ding{51}     &  49.24 & 39.03 \\ 
  \hline
\end{tabular}
}
\label{ablation_prompt}
\end{table}

Table \ref{ablation_prompt} presents the evaluation results of utilizing different prompts. The point prompt delivers the worst performance and negatively affects the accuracy of any prompt combinations. This could be attributed to the insufficient amount of foreground points, which cannot guide the model effectively. The mask prompt performs better than the point prompt, but it still cannot generate high-quality segmentation annotations. The highest accuracy achieved by a mask prompt is approximately 60\%, which is still much lower than the optimal prompts. Furthermore, the mask prompt has a negative impact on the performance of box prompts. When solely adopting the H-Box prompt, we obtain the highest accuracy compared to the point and mask prompts. For the case of utilizing R-Box annotations, the RH-Box prompt also achieves satisfactory performance. From this experiment, we conclude that: \textit{if an RS object detection dataset only has R-Box annotations, then the RH-Box prompt should be used; otherwise, the H-Box prompt should be adopted.} This consideration is applied in our later dataset transformations.

\begin{table*}[t]
  \caption{Comparisons of different high-resolution RS segmentation datasets.}
  \newcommand{\tabincell}[2]{\begin{tabular}{@{}#1@{}}#2\end{tabular}}
  \centering
  \begin{threeparttable}
  \resizebox{\linewidth}{!}{
    \begin{tabular}{lccccccc}
  \hline
  Dataset & \#Images & \#Category & \#Channels & Resolution (m)  & Image size & Instance  & Fine-grained \\
  \hline
  ISPRS Vaihingen \footnotemark[1] & 33 & 6 & IR,R,G & 0.09 & 2,494 $\times$ 2,064 & & \\
  ISPRS Potsdam \footnotemark[5] & 38 & 6 & IR,RGB & 0.05 & 6,000 $\times$ 6,000 & &\\
  Zurich Summer \cite{zurich_summer} & 20 & 8 & NIR,RGB & 0.62 & 1,000 $\times$ 1,150 &  &\\
  Zeebruges \cite{zeebruge} & 7 & 8  & RGB & 0.05 & 10,000 $\times$ 10,000 & &\\
  DeepGlobe Land Cover \cite{deepglobe} & 1,146 & 7 & RGB & 0.5 & 2,448 $\times$ 2,448 &  &\\
  UAVid \cite{uavid} & 420 & 8 & RGB & - & 4,096 $\times$ 2,160 or 3,840 $\times$ 2,160 &  & \\
  GID \cite{gid} & 150 & 15 & NIR,RGB & 1 or 4  & 6,800 $\times$ 7,200 & &\\ 
  Landcover.ai \cite{landcover} & 41 & 3 & RGB & 0.25 or 0.5 & 9,000 $\times$ 9,500 or 4,200 $\times$ 4,700 &  & \\
  IsAID \cite{isaid} & 2,806 & 15 & RGB &  - & 800 $\times$ 800 $\sim$ 4,000 $\times$ 13,000 & \ding{51} & \\
  LoveDA \cite{loveda} & 5,987  & 7 &  RGB & 0.3 & 1,024 $\times$ 1,024 & & \\  
  \hline
  \hline
  \bfseries {SAMRS} & \multicolumn{7}{c}{}\\
  \hline
  \textbf{SOTA} & 17,480 & 18 & RGB & - & 1,024 $\times$ 1,024  & \ding{51} & \\
  \textbf{SIOR} & 23,463 \tnote{1} & 20 & RGB & - & 800 $\times$ 800 & \ding{51}  & \\
  \textbf{FAST} & 64,147 & 37 & RGB & - & 600 $\times$ 600 & \ding{51} & \ding{51} \\
  \hline
\end{tabular}
}
\begin{tablenotes}
\scriptsize
\item[1] To avoid data snooping, only the 11725 images corresponding to the original DIOR trainval dataset are used in subsequent pre-trainings.
\end{tablenotes}
\end{threeparttable}
\label{compare}
\end{table*}

\footnotetext[5]{\url{https://www.isprs.org/education/benchmarks/UrbanSemLab/2d-sem-label-potsdam.aspx}}

\subsection{Dataset Transformation} 

For the FAIR1M-2.0 dataset, since it only contains R-Box annotations, we use the corresponding RH-Box as the prompt. For DOTA-V2.0 and DIOR, we directly adopt the H-Box prompt. Prior to transformation, we follow the common practice to crop images in DOTA and FAIR1M datasets to 1,024 $\times$ 1,024 and 600 $\times$ 600, respectively, while images in DIOR are maintained at the size of 800 $\times$ 800. The resulting datasets are named SOTA (\textit{i.e.}, \textcolor{blue}{D}OTA $\rightarrow$ \textcolor{red}{S}OTA), SIOR (\textit{i.e.}, \textcolor{blue}{D}IOR $\rightarrow$ \textcolor{red}{S}IOR), and FAST (\textit{i.e.}, \textcolor{blue}{F}ine-gr\textcolor{blue}{A}ined object recogn\textcolor{blue}{I}tion in high-\textcolor{blue}{R}esolution remote sensing imagery $\rightarrow$ \textcolor{red}{F}ine-gr\textcolor{red}{A}ined \textcolor{red}{S}egmentation for high-resolution remo\textcolor{red}{T}e sensing imagery), respectively. These datasets constitute a comprehensive and large-scale remote sensing segmentation database, \textit{i.e.}, \textbf{SAMRS}.

\section{SAMRS}
\subsection{Basic Information}

The obtained segmentation labels are stored in *.png files. Pixel values are aligned with the object classes of source object detection datasets. The areas that have not been covered by the generated masks will be in a pixel value of 255. We present the comparison of our SAMRS dataset with existing high-resolution RS segmentation datasets in Table \ref{compare} from different aspects. With the available high-resolution RSI object detection datasets, we can efficiently annotate 105,090 images containing 1,668,241 instances based on SAM and the identified prompt settings (Sec.~\ref{subsec:ablation}), which is more than ten times the capacity of existing datasets. Additionally, SAMRS inherits the categories of the original detection datasets, which makes them more diverse than other high-resolution RS segmentation collections. It is worth noting that RS object datasets usually have more diverse categories than RS segmentation datasets due to the difficulty of tagging pixels in RSIs, and thus our SAMRS reduces this gap. 

Specifically, the resulting FAST dataset is a large-scale fine-grained RS segmentation dataset that targets diverse vehicles and grounds, while SOTA and SIOR are segmentation datasets containing common object categories. For this reason, we did not unify their categories. In addition to the massive pixel-level semantic mask annotations, SAMRS includes instance mask and bounding box annotations. This means that \textit{it can be used to perform semantic segmentation, instance segmentation, and object detection, either individually or in combination.} This feature sets SAMRS apart from the IsAID dataset, which was independently annotated from scratch on DOTA-V1.0 \cite{dota} images.

\begin{figure*}[t]
    \centering
    \includegraphics[width=0.9\linewidth]{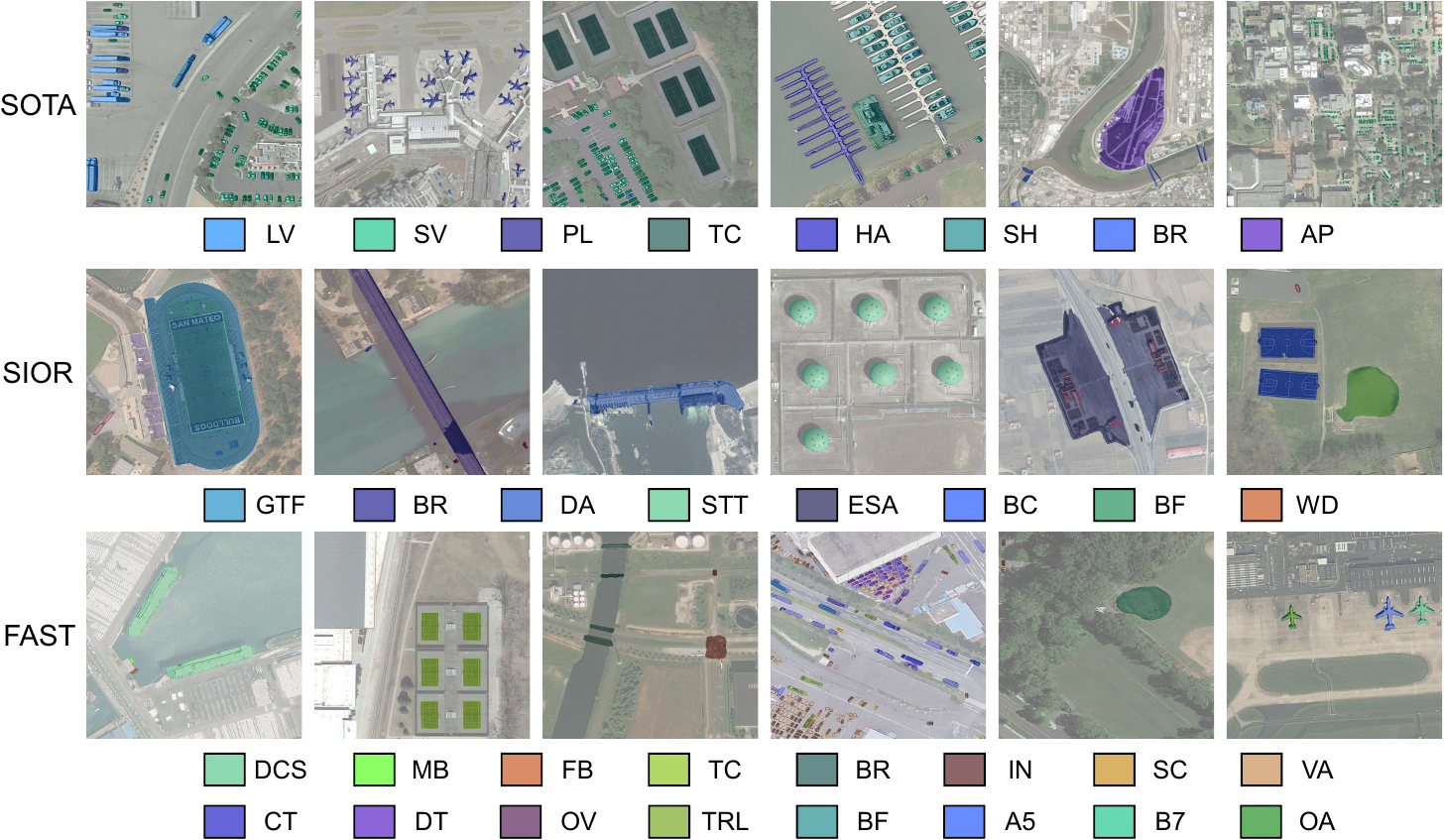}
    \caption{Some visual examples from the three subsets of our SAMRS dataset. For the definition of classes, please refer to the supplementary material.}
    \label{seg_annots}
\end{figure*}

\subsection{Statistics and Analysis}

\begin{figure*}[h]
    \centering
    \includegraphics[width=\linewidth]{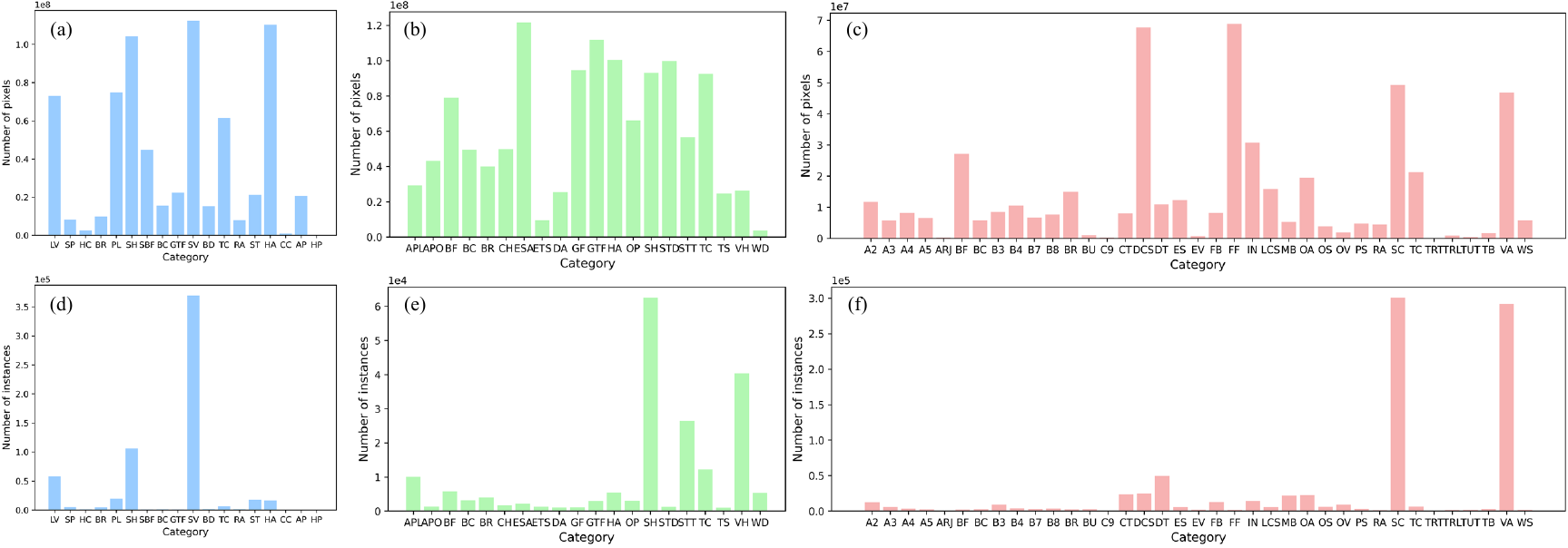}
    \caption{Statistics of the number of pixels and instances per category in SAMRS. The histograms for the subsets SOTA, SIOR, and FAST are shown in the first, second, and third columns, respectively. The first row presents histograms on a per-pixel basis, while the second row presents histograms on a per-instance basis. A list of category abbreviations is provided in the supplementary material.}
    \label{class_pixel_ins}
\end{figure*}

\begin{figure*}[h]
    \centering
    \includegraphics[width=\linewidth]{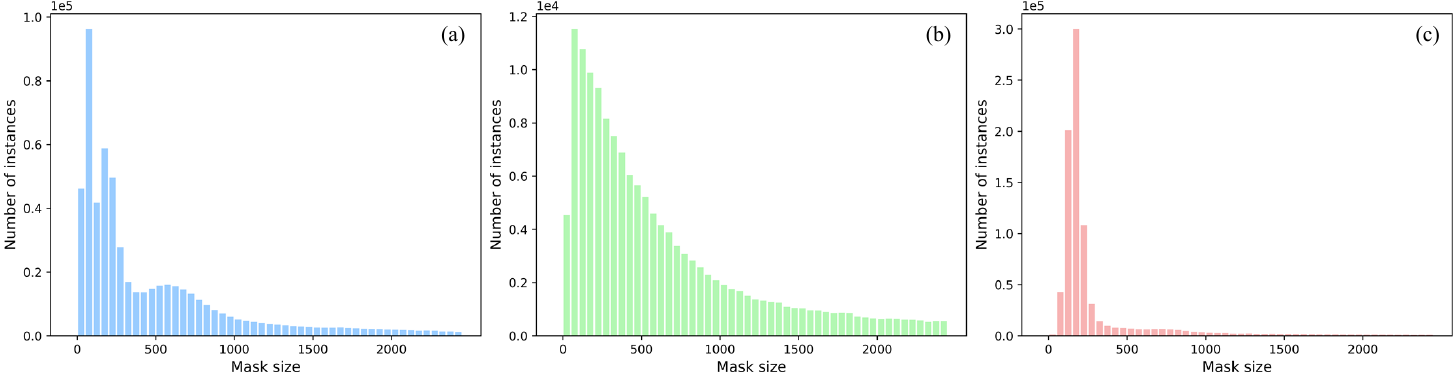}
    \caption{Statistics of the mask sizes in the subsets of SAMRS. (a) SOTA. (b) SIOR. (c) FAST.}
    \label{mask_size}
\end{figure*}

To gain a deeper understanding of the characteristics of the SAMRS dataset, we conduct a thorough analysis of their capacity per category, including pixel and instance numbers. The results are presented in Figure \ref{class_pixel_ins}. In this analysis, we only count instances that have valid masks. The figure indicates that SIOR has more balanced categories compared to SOTA and FAST. In the instance-level statistics, we observe a large number of vehicle annotations, particularly on small ships and cars, as they are common in the real world and frequently appear in RSIs. This could also be the goal of initially developing these detection datasets. For instance, DOTA-V2.0 focuses on small targets, while FAIR1M mainly aims to accurately distinguish between different types of vehicles. Furthermore, it is observed that some categories have a high number of pixels but a low number of instances, which is likely due to their large size. For instance, the \textit{expressway-service-area} in SIOR and the \textit{football-field} in FAST demonstrate this pattern.

In addition, we investigate the distribution of mask sizes in SAMRS, as shown in Figure \ref{mask_size}. The results indicate that, in general, there are more instances with smaller sizes in all subsets. However, some differences exist between the subsets. Specifically, FAST has more small objects than the other two sets. Nevertheless, SOTA appears to have a higher number of extremely small targets (\textit{i.e.}, \textless 100 pixels), since its source dataset DOTA-V2.0 is designed for small object detection. On the other hand, SIOR has a more smooth distribution of mask sizes compared to SOTA and FAST.

\subsection{Visualization} 

In Figure \ref{seg_annots}, we visualize some segmentation annotations from the three subsets in our SAMRS dataset. As can be seen, SOTA exhibits a greater number of instances for tiny cars, whereas FAST provides a more fine-grained annotation of existing categories in SOTA such as car, ship, and plane. SIOR on the other hand, offers annotations for more diverse ground objects, such as \textit{dam}. Hence, our SAMRS dataset encompasses a wide range of categories with varying sizes and distributions, thereby presenting a new challenge for RS semantic segmentation.

\section{Experiment} 

\subsection{Pre-training}

\subsubsection{Data and Model Settings}

To investigate the influence of segmentation pre-training (SEP) using SAMRS, we adopt multiple segmentation frameworks, including typical encoder-decoder networks and the recently emerged end-to-end structure. In encoder-decoder networks, we utilize classical UNet \cite{unet} and commonly-used UperNet \cite{upernet}. Different from the original U-Net that has five blocks in the decoder part, to be compatible with the typical hierarchical pyramid backbone network that outputs four levels of features, we replace the last block to a single 2$\times$ bilinear upsampling layer, which is followed by a segmentation head that contains a 1$\times$1 convolution, a 2$\times$ bilinear upsampling, and a ReLU activation function. For UperNet, the segmentation head only employs a 1$\times$1 convolution. To comprehensively explore the SEP, in addition to traditional convolutional networks such as ResNet \cite{resnet}, diverse backbones are used, including hierarchical vision transformers: Swin \cite{swint}, ViTAEv2 \cite{vitae_v2} and InternImage \cite{internimage}, and non-hierarchical networks, including ViT \cite{vit}, ViT-Adapter \cite{vit_adapter} and ViT-RVSA \cite{rvsa}. For the end-to-end structure, we choose the recent Mask2Former \cite{mask2former}. The SAMRS is split into two parts, one for pre-training, and another for validation, see the supplementary material for more details. In the data preprocessing, we employed common data augmentation techniques, including random scaling, random horizontal and vertical flipping, random rotation by 90 degrees, and altering pixel values through random color jitter and gamma transformation. Moreover, to ensure a fair comparison with prior studies \cite{imagenet, rsp}, we randomly cropped the input images to a size of 224 $\times$ 224.

\begin{figure}[t]
    \centering
    \includegraphics[width=\linewidth]{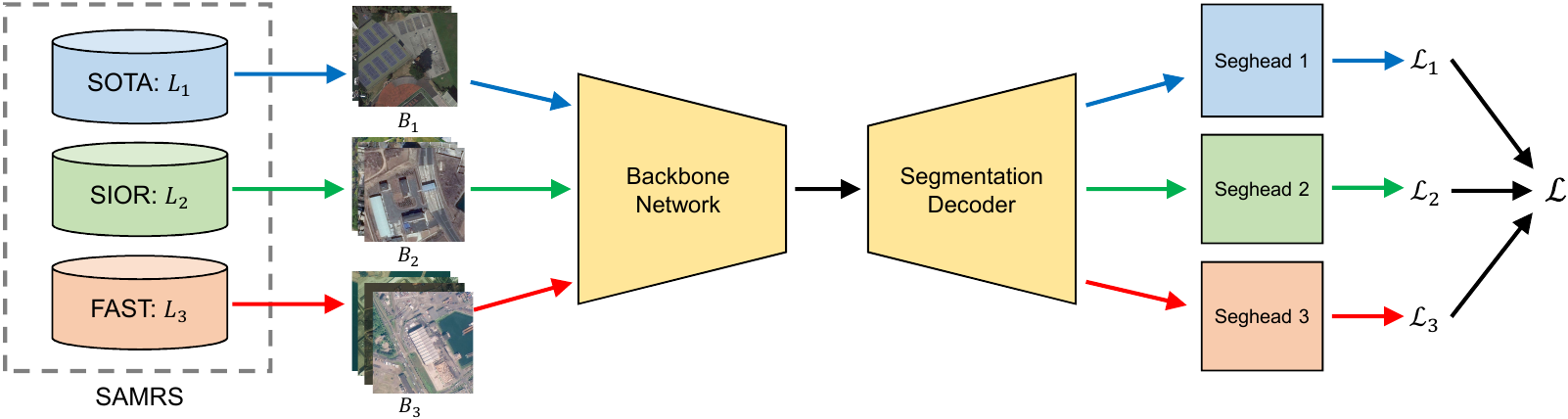}
    \caption{The pipeline of segmentation pre-training on SAMRS. Different colors represent the data stream of various sets. The yellow parts will be used in fine-tuning.}
    \label{seg_pretrain}
\end{figure}

\subsubsection{Training Settings}
\label{sec:train_setting}

Intuitively, a well-initialized backbone network generates discriminative features at the beginning of training, thereby facilitating the optimization process of the decoder component. Since the SEP is expected to mitigate the gaps between pre-training and downstream tasks. To this end, before the segmentation pre-training phase, the selected model's backbone network is initialized using pretrained weights. In our experiments, to fully evaluate the SEP, besides basic supervised pre-training on ImageNet (IMP) \cite{imagenet}, we also utilize the RSP \cite{rsp} on the MillionAID Dataset \cite{millionaid}. In addition, the unsupervised pre-training weights are also involved, including BEiT \cite{beit} and MAE \cite{mae}. Here, the MAE pre-training is conducted on the MillionAID \cite{rvsa}, while the BEiT is pretrained on the ImageNet.

To accommodate the multiple segmentation sets within SAMRS, each having a different number of categories, we employ a multi-head pre-training strategy. This approach involves utilizing separate segmentation heads for individual datasets. The only distinction lies in the output channel count of the 1 $\times$ 1 convolution, which corresponds to the number of categories. During batch-based training, diverse mini-batches are sampled from these sets to form a collective large batch, which is then fed into the network. Given the volume disparities among the various sets in SAMRS, proportional sampling is employed to obtain the mini-batches. Assuming a large batch of size $B$ consists of $M$ mini-batches with sizes $B_1, B_2, \cdots, B_M$, it follows that $B = B_1 + B_2 + \cdots + B_M$. Each mini-batch corresponds to its respective segmentation head, resulting in a training loss $\mathcal{L}_i$ for the $i$th mini-batch. The total loss is computed as $\mathcal{L} = \mathcal{L}_1 + \mathcal{L}_2 + \cdots + \mathcal{L}_M$. Assuming the sizes of the $M$ sets in SAMRS are $L_1, L_2, \cdots, L_M$, we can express this relationship as $B_i = \frac{L_i}{\sum_{j=1}^M L_j} B, \ i=0,1,...,M$. Here, $M$ can be easily extended for more RS detection datasets. In this study, $M=3$. Figure~\ref{seg_pretrain} illustrates the pre-training pipeline. Each model is first pre-trained for 80k iterations with $B=96$ and then used for fine-tuning. All experiments are implemented by PyTorch on NVIDIA GeForce RTX 3090 GPUs.

\begin{table}[t]
  \caption{Segmentation results of different methods on the ISPRS Potsdam dataset. $\ddagger$: MAE pre-training on the MillionAID. $\ddagger\ddagger$: MAE pre-training on the SAMRS training set. ``*'' denotes the best score among all methods.}
  \centering
  \resizebox{\linewidth}{!}{
    \begin{tabular}{l|l|l|ccccc|cc}
  \hline
    \multirow{2}*{Method} & \multirow{2}*{Pretrain} & \multirow{2}*{Backbone} & \multicolumn{5}{c|}{F1 score per category} & \multirow{2}*{OA} & \multirow{2}*{mF1} \\
  \cline{4-8}
  & & & Imper. surf. & Building & Low veg. & Tree & Car & &  \\
  \hline
  \multicolumn{3}{l|}{\bfseries \textit{Comparison method}} & \multicolumn{6}{c}{} \\
  \hline
  ST-UNet \cite{stunet}  & --- & ResNet-50 & 79.19  & 86.63  & 67.89 & 66.37 & 79.77  & --- & 86.13 \\
  ResUNet-a d7v2 \cite{resunet}  & --- & --- & 93.50 &  97.20 & 88.20 & 89.20 & 96.40 & 91.50 &  92.90 \\
  LANet \cite{lanet} & IMP & ResNet-50 & 93.05 & 97.19  & 87.30 & 88.04 & 94.19 & 90.84 & 91.95  \\
  DCFAM \cite{dc_swin} & IMP & Swin-S & 94.19 & 97.57  & 88.57 & 89.62 & 96.31 & 92.00 & 93.25* \\
  \hline
  \multicolumn{3}{l|}{\bfseries \textit{Convolutional network}} & \multicolumn{6}{c}{} \\
  \hline
  UNet & SEP & ResNet-50 & 90.62 & 94.75 & 85.12 & 83.91 & 96.51 & 89.70 & 90.18 \\
  UNet & IMP & ResNet-50 & 90.78 & 94.78 & 85.23 & 84.76  & 96.81  & 89.94 & 90.47 \\
  UNet & IMP+SEP & ResNet-50 & 91.36 & 94.92 & 85.39 & 85.24 & 97.17  & \bfseries 90.29 & \bfseries 90.82 \\
  \hline
  UperNet & SEP & ResNet-50 & 91.02  & 94.82 & 84.28 & 83.97 &  96.95 & 89.70 & 90.21 \\
  UperNet & IMP & ResNet-50 & 90.70  & 94.44  & 84.68 & 83.94 & 96.58 & 89.59 & 90.07 \\
  UperNet & IMP+SEP & ResNet-50 & 91.38 & 95.26  & 85.14 & 84.88 & 97.16 & \bfseries 90.27 & \bfseries 90.76 \\
  \hline
  \multicolumn{3}{l|}{\bfseries \textit{Hierarchical vision transformer}} & \multicolumn{6}{c}{} \\
  \hline
  UperNet & IMP & Swin-T &  93.09 & 96.74  & 86.99 & 86.45 & 91.12 & 91.44 & 90.88 \\
  UperNet & IMP+SEP & Swin-T & 93.06 &  96.65 & 87.07 & 86.74 & 97.64 & \bfseries 91.88  & \bfseries 92.23 \\
  \hline
  UperNet & IMP & ViTAEv2-S & 92.54  &  96.54 & 86.11 & 86.13 & 91.31 & 91.00 & 90.52 \\
  UperNet & IMP+SEP & ViTAEv2-S & 93.45 &  96.99 & 87.65 & 87.00 & 97.67 & \bfseries 92.25*  & \bfseries 92.55 \\
  \hline
  UperNet & IMP & InternImage-T &  93.27 & 96.80  & 87.41 & 86.62 & 91.79 & 91.65 & 91.18 \\
  UperNet & IMP+SEP & InternImage-T & 93.30 & 96.91  & 87.24 & 86.80 & 97.81 & \bfseries 92.08 & \bfseries 92.41 \\
  \hline
  \multicolumn{3}{l|}{\bfseries \textit{Plain vision transformer}} & \multicolumn{6}{c}{} \\
  \hline
  UperNet & IMP & ViT-B & 93.09  &  96.83 & 86.93 & 86.61 & 90.93 & 91.47 & 90.88 \\
  UperNet & IMP+SEP & ViT-B & 92.96 &  96.52 & 86.62 & 86.01 & 97.57 & \bfseries 91.60 & \bfseries 91.94 \\
  \hline
  UperNet & IMP & ViT-Adapter-B &  93.16 & 96.77  & 87.09 & 86.71 & 91.20 & 91.53 & 90.98 \\
  UperNet & IMP+SEP & ViT-Adapter-B & 93.20 &  96.75 & 87.06 & 86.52 & 97.68 & \bfseries 91.91  & \bfseries 92.24 \\
  \hline
  \multicolumn{3}{l|}{\bfseries \textit{Pre-training strategy}} & \multicolumn{6}{c}{} \\
  \hline
  UNet & RSP & ResNet-50 & 91.49 & 95.42 & 85.70 & 85.18 & 97.05  & 90.49 & 90.97 \\
  UNet & RSP+SEP & ResNet-50 & 92.00 & 95.44 & 85.76  & 85.33 & 97.38 & \bfseries 90.72 & \bfseries 91.18 \\
  \hline
  UperNet & RSP & ResNet-50 & 91.08 & 94.64 & 85.57 & 85.38 & 96.97 & 90.18 &  90.73 \\
  UperNet & RSP+SEP & ResNet-50 & 91.73 &  95.52 & 85.44 & 85.35 & 97.24 & \bfseries 90.59 & \bfseries 91.06 \\
  \hline
  UperNet & BEiT & ViT-B &  88.70 & 92.29  & 81.48 & 78.64 & 96.36 &  86.86 & 87.49 \\
  UperNet & BEiT+SEP & ViT-B & 89.95 & 93.33  & 82.96 & 80.91 & 96.67 & \bfseries 88.20 & \bfseries 88.76 \\
  \hline
  UperNet & MAE $\ddagger$ & ViT-B + RVSA & 92.67  & 96.38  & 86.43 & 85.89 & 90.46 & 90.97 & 90.37 \\
  UperNet & MAE+SEP & ViT-B + RVSA & 92.69 &  96.33 & 86.28 & 85.60 & 97.56 & \bfseries 91.33 & \bfseries 91.69 \\
  \hline
  UperNet & SAMRS-MAE $\ddagger\ddagger$ & ViT-B + RVSA &  92.46 & 96.10 & 86.18 & 85.59 & 90.35 & 90.71  & 90.13  \\
  UperNet & SAMRS-MAE+SEP & ViT-B + RVSA  & 92.34 & 95.88 & 86.06 & 85.32 & 97.54 & \bfseries 91.01 & \bfseries 91.43 \\
  \hline
  \multicolumn{3}{l|}{\bfseries \textit{End-to-end transformer}} & \multicolumn{6}{c}{} \\
  \hline
  Mask2Former & IMP & ResNet-50 & 88.40  & 92.93  & 83.05 & 83.98 & 86.00 & \bfseries 87.54 & \bfseries 86.87 \\
  Mask2Former & IMP+SEP & ResNet-50 & 72.41 &  78.98 & 63.14 & 61.62 & 73.16 &  70.14 & 69.86 \\
  \hline
\end{tabular}
}
\label{compare_potsdam}
\end{table}

\subsection{Fine-tuning}

\subsubsection{Comparison to Various Pre-training Strategies}

In the RS community, ISPRS Potsdam and IsAID are commonly-used finely annotated datasets for evaluating segmentation methods \cite{rsp, rvsa, hmanet, rssformer,ass_2022_tgrs_factseg}, and we use them to assess the pre-trained models, as Table \ref{compare_potsdam}-\ref{compare_isaid} shown. It can be seen that, without good initialization, the performances of SEP are comparable as IMP but inferior to IMP+SEP. On the Potsdam scene, for the traditional encoder-decoder network, SEP improves both convolutional and vision transformer networks, especially for UperNet and the backbones containing hierarchical features (also includes ResNet). As a result, ViTAEv2-S is greatly boosted and overperforms existing advanced methods in overall accuracy. We also observe that SEP is useful when combined with different pre-training strategies. Even if using the initialized weights generated by pre-training on SAMRS itself, SEP still can improve the accuracy, excluding the effect of data volume used for training. We notice SEP played a negative role in the end-to-end structure, it may be because the objects of SAMRS are too small, which is unfavorable for the region-based Mask2Former. In addition, the Mask2Former, which has obtained high accuracies on natural images, does not perform as well as UNet and UperNet on RSIs. These results indicate more refined parameter adjustments of Mask2Former are needed in later research. On the IsAID dataset, the performances of SEP on simple convolutional networks depending on local perception are unstable, because IsAID and DOTA share the same images but with different annotations, which may confuse the model. Benefiting from SEP, vision transformer networks are further enhanced and surpass previous methods. From these results we can see, SEP is able to mitigate the influence of task-level disparities, specifically the gaps between upstream pre-training tasks and downstream segmentation tasks.

\begin{table*}[t]
  \caption{Segmentation results of different methods on the IsAID dataset. $\ddagger$: MAE pre-training on the MillionAID. ``*'' denotes the best score among all methods.}
  \newcommand{\tabincell}[2]{\begin{tabular}{@{}#1@{}}#2\end{tabular}}
  \centering
  \begin{threeparttable}
  \resizebox{\linewidth}{!}{
  \begin{tabular}{l|l|l|ccccccccccccccc|c}
  \hline
  \multirow{2}*{Method} & \multirow{2}*{Pretrain} & \multirow{2}*{Backbone} & \multicolumn{15}{c|}{IOU per category\tnote{1}} & \multirow{2}*{mIOU}  \\
  \cline{4-18}
  & & & SH & ST & BD & TC & BC & GTF & BR & LV & SV & HC & SP & RA & SBF & PL & HA &  \\
  \hline
  \multicolumn{3}{l|}{\bfseries \textit{Comparison method}} & \multicolumn{16}{c}{} \\
  \hline
  HMANet \cite{hmanet} & IMP & ResNet-50  & 65.38 & 70.92 & 74.71 & 88.69 & 60.51 & 54.57 & 28.98 & 59.74 & 50.28 & 32.58 & 51.41 & 62.88 & 70.20 & 83.79 & 51.91 & 62.64 \\
  UperNet & MAE $\ddagger$ & ViTAE-B + RVSA$^\Diamond$ & 70.61 & 77.19 & 73.14 & 70.53 & 53.46 & 59.69 & 44.78 & 63.89 & 52.56 & 38.05 & 47.07 & 66.51 & 74.61 & 85.78 & 54.90 & 64.49 \\
  FactSeg \cite{ass_2022_tgrs_factseg}  & IMP & ResNet-50  & 68.34 & 56.83 & 78.36 & 88.91 &  64.89 & 54.60 & 36.34 & 62.65 & 49.53 & 42.72 & 51.47 & 69.42 & 73.55 & 84.13 & 55.74 & 64.79 \\
  RssFormer \cite{rssformer} & IMP & RSS-B & --- & --- & ---& --- & --- & --- & --- & --- & ---& --- & --- & --- & --- & --- & --- & 65.88 \\
  \hline
  \multicolumn{3}{l|}{\bfseries \textit{Convolutional network}} & \multicolumn{16}{c}{} \\
  \hline
  UNet & IMP & ResNet-50 &  50.11 & 48.50 & 31.52 & 74.30 & 15.38 & 14.99 & 8.83 & 51.59  & 34.71 & 0.07  & 39.03 & 7.88 &  39.33 & 73.12  & 47.71  & 35.80 \\
  UNet & IMP+SEP & ResNet-50 &  62.01 & 62.58 & 66.26 & 85.84 & 51.65 & 37.77 & 31.99 & 61.03  & 45.57 & 3.14  & 45.61 & 56.29 & 64.40  & 81.58  &  54.22 & \bfseries 54.00 \\
  \hline
  UNet & RSP & ResNet-50 & 53.00 &  53.53 & 62.99 & 78.25 & 34.48 & 36.92  & 24.83  & 54.95 & 34.22 & 2.11 & 40.36 & 42.18 & 50.37 & 77.82 & 48.81 & \bfseries 46.32 \\
  UNet & RSP+SEP & ResNet-50 & 46.83 & 28.82  & 26.58 & 67.30 & 13.83 & 17.79  & 3.97  & 48.37 & 28.30 & 0.00 & 35.30 & 0.00 & 28.74 & 67.62 & 43.29 & 30.45 \\
  \hline
  UperNet & IMP & ResNet-50 &  69.81 & 71.08 & 75.99 & 87.73 & 58.11 & 61.96 & 40.06 & 63.84  & 48.54 & 27.02  & 48.59 & 70.19 & 76.15  & 82.94  & 56.05  & \bfseries 62.54 \\
  UperNet & IMP+SEP & ResNet-50 & 66.37 & 65.80 &  70.24 & 86.60 & 54.35 & 47.23 & 31.71 & 63.46  & 46.25  & 22.26  & 48.01 & 61.29 & 72.28  & 81.22  & 54.44  & 58.10 \\
  \hline
  UperNet & RSP & ResNet-50 & 46.76 & 13.21 & 35.21 & 56.86 & 13.89 & 22.78  & 3.91  & 47.44 & 24.10 & 0.32 & 40.26 & 27.55 &48.42  & 68.92 & 44.88 & 32.97 \\
  UperNet & RSP+SEP & ResNet-50 & 66.62 & 65.37  & 72.49 & 86.88 & 57.18 & 50.20  & 37.71  & 63.29 & 47.41 & 22.96 & 45.36 & 65.98 & 74.91 & 77.97 & 54.53 & \bfseries 59.26 \\
  \hline
  \multicolumn{3}{l|}{\bfseries \textit{Vision transformer}} & \multicolumn{16}{c}{} \\
  \hline
  UperNet & IMP & Swin-T  & 71.26 & 73.97  & 77.54 & 87.33 & 56.42 & 61.53  & 42.36  & 65.89 & 49.74 & 38.30 & 47.82 & 66.08 & 76.22 & 84.72 & 57.39 & 63.77 \\
  UperNet & IMP+SEP & Swin-T  & 73.37 & 74.19  & 76.63 & 87.63 & 56.40 & 61.44  & 42.42  & 65.98 & 50.92 & 38.82 & 49.16 & 69.70 & 75.05 & 84.83 & 57.27 & \bfseries 64.25 \\
  \hline
  UperNet & IMP & ViTAEv2-S  & 73.31 & 76.22  & 75.12 & 89.65 & 62.72 & 65.71  & 42.94  & 67.09 & 52.01 & 39.56 & 48.07 & 71.38 & 79.04 & 85.64 & 58.47 & 65.79 \\
  UperNet & IMP+SEP & ViTAEv2-S  & 75.95 &  75.48 & 75.38 & 89.56 & 63.79 & 64.85  & 45.62  & 68.49 & 53.27 & 37.56 & 49.49 & 69.84 & 79.22 & 85.85 & 59.58 & \bfseries 66.26* \\
  \hline
  UperNet & IMP & InternImage-T  & 74.56 &  73.71 & 79.04 & 88.04 & 65.11 & 64.90  & 38.35  & 65.53 & 51.16 & 39.00 & 48.79 & 66.35 & 76.13 & 85.39 & 59.25 & 65.02 \\
  UperNet & IMP+SEP & InternImage-T & 75.12 & 73.67  & 79.34 & 88.58 & 62.39 & 65.37  & 42.54  & 67.35 & 52.91 & 42.09 & 48.21 & 69.94 & 79.33 & 86.09 & 57.88 & \bfseries 66.05 \\
  \hline
\end{tabular}
  }
  \begin{tablenotes}
    \scriptsize
    \item[1] SH: ship. ST: storage tank. BD: baseball diamond. TC: tennis court. BC: baseball court. GTF: ground track field. BR: bridge. LV: large \\ vehicle. SV: small vehicle. HC: helicopter. SP: swimming pool. RA: roundabout. SBF: soccer ball field. PL: plane. HA: harbor.
  \end{tablenotes}
  \end{threeparttable}
  \label{compare_isaid}
\end{table*}

\subsection{Fine-tuning with Small-size Training Samples}

\begin{table}[h]
\begin{minipage}{0.48\linewidth}
  \caption{Segmentation results of different pre-training methods on the ISPRS Potsdam dataset.}
  \centering
\resizebox{\linewidth}{!}{
    \begin{tabular}{c|l|c|ccc}
  \hline
    \multirow{2}*{Method} & \multirow{2}*{Pretrain} & \multirow{2}*{Backbone} & \multicolumn{3}{c}{mF1} \\
  \cline{4-6}
  & & & 5\% & 3\% & 1\%  \\
  \hline
 UNet & IMP & ResNet-50 & 80.45 & 75.03 & 65.72 \\
UNet & IMP+SEP & ResNet-50 & 81.80 & 77.97 & 69.70 \\
&  &  $\Delta$ & +1.35 & +2.94 & +3.98 \\
\hline
UNet & RSP & ResNet-50 & 80.34 & 75.35 & 62.86 \\
 UNet & RSP+SEP & ResNet-50 & 81.69 & 77.40 & 68.35 \\
 &  &  $\Delta$ & +1.35 & +2.05 & +5.49 \\
  \hline
\end{tabular}
}
\label{data_efficient_potsdam}
\end{minipage}
\hspace{2mm}
\begin{minipage}{0.48\linewidth}
  \caption{Segmentation results of different pre-training methods on the IsAID dataset.}
  \centering
  \resizebox{\linewidth}{!}{
    \begin{tabular}{c|l|c|ccc}
  \hline
    \multirow{2}*{Method} & \multirow{2}*{Pretrain} & \multirow{2}*{Backbone} & \multicolumn{3}{c}{mIOU} \\
  \cline{4-6}
  & & & 5\% & 3\% & 1\%  \\
  \hline
 UNet & IMP & ResNet-50 & 5.33 & 5.19 & 1.33 \\
UNet & IMP+SEP & ResNet-50 & 18.74 & 12.98 & 7.04 \\
&  &  $\Delta$ & +13.41 & +7.79 & +5.71 \\
\hline
UNet & RSP & ResNet-50 & 4.68 & 3.67 & 2.50 \\
 UNet & RSP+SEP & ResNet-50 &22.47 & 13.88 & 8.84 \\
 &  &  $\Delta$ & +17.79 & +10.21 & +6.34 \\
  \hline
\end{tabular}
}
\label{data_efficient_isaid}
\end{minipage}
\end{table}

The difficulty of annotating pixel-level masks limits the scale of existing remote sensing segmentation datasets, ultimately constraining the performance of trained models due to insufficient training samples. In order to investigate the effectiveness of SEP under conditions of limited training samples, we conducted experiments involving fine-tuning models using small fractions (1\%, 3\%, and 5\%) of data from the ISPRS Potsdam and IsAID training sets, as outlined in Table~\ref{data_efficient_potsdam} and Table~\ref{data_efficient_isaid}. The integration of SEP with SAMRS, which provides a valuable segmentation prior, yields superior results compared to the IMP and RSP counterparts. This advantage is particularly evident when the number of available samples is scarce in the ISPRS Potsdam scene. Conversely, the results for the IsAID dataset exhibit an opposite trend due to the inherent challenges of this dataset, where both IMP and RSP yield extremely low overall accuracies. Nevertheless, the adoption of SEP significantly improves model performance. These findings highlight the importance of conducting segmentation pre-training using large-scale RS segmentation data, such as SAMRS, prior to training with limited data. Notably, the developed pipeline enables the rapid construction of such a dataset at a low labeling cost, making it a promising approach.

\subsection{Limitations and Discussion}
\label{sec:limit}

Previous classical datasets, such as HRSC2016 \cite{hrsc2016} and COCO \cite{coco}, simultaneously contain bounding box and pixel-level mask annotations, proving the feasibility of the coexistence for segmentation and detection labels. Therefore, it is reasonable to construct the SAMRS dataset by transforming existing RS object detection datasets. However, despite successfully establishing the SAMRS dataset, which outperforms existing high-resolution RS segmentation datasets by more than tenfold, its volume remains smaller than large-scale classification datasets such as ImageNet \cite{imagenet} and MillionAID \cite{millionaid}, commonly employed for pre-training purposes. Our current investigation focuses exclusively on pre-training small-scale basic models (about 100M) and we intend to incorporate larger models. Additionally, it is also worth trying to explore the impact of pre-training on SAMRS for tasks such as instance segmentation and object detection.

\section{Conclusion}

This study presents an effective way to create a large-scale remote sensing (RS) segmentation dataset by harnessing the capabilities of the Segment Anything Model (SAM) and existing object detection datasets. Given the unique challenges associated with RS data labeling, we investigate the performance of various prompts to identify the optimal settings for SAM. By leveraging these optimal settings, we generate extensive mask annotations for RS images, thereby creating a large-scale segmentation dataset named SAMRS. Remarkably, SAMRS surpasses all previously available high-resolution RS segmentation datasets in terms of volume. Furthermore, our statistical analysis reveals that SAMRS encompasses a diverse array of categories exhibiting varying sizes and distributions. SAMRS can be utilized for semantic segmentation, instance segmentation, and object detection, either independently or in combination. Specifically, we present a preliminary investigation and demonstrate the value of segmentation pre-training on SAMRS for RS segmentation tasks, especially in scenarios with limited training samples.

\section*{Acknowledgments}

We acknowledge the authors of SAM for releasing codes and models, and the authors of DOTA, DIOR, and FAIR1M for providing their datasets. This work was supported in part by the National Natural Science Foundation of China under Grant 62225113 and in part by the National Key Research and Development Program of China under Grant 2022YFB3903405.


\bibliographystyle{ieee_fullname}
\bibliography{egbib}

\begin{thebibliography}{10}\itemsep=-1pt

\bibitem{beit}
Hangbo Bao, Li Dong, Songhao Piao, and Furu Wei.
\newblock {BE}i{T}: {BERT} pre-training of image transformers.
\newblock In {\em ICLR}, 2022.

\bibitem{landcover}
Adrian Boguszewski, Dominik Batorski, Natalia Ziemba-Jankowska, Tomasz
  Dziedzic, and Anna Zambrzycka.
\newblock Landcover. ai: Dataset for automatic mapping of buildings, woodlands,
  water and roads from aerial imagery.
\newblock In {\em CVPR}, pages 1102--1110, 2021.

\bibitem{rsprompter}
Keyan Chen, Chenyang Liu, Hao Chen, Haotian Zhang, Wenyuan Li, Zhengxia Zou,
  and Zhenwei Shi.
\newblock Rsprompter: Learning to prompt for remote sensing instance
  segmentation based on visual foundation model.
\newblock {\em arXiv preprint arXiv:2306.16269}, 2023.

\bibitem{vit_adapter}
Zhe Chen, Yuchen Duan, Wenhai Wang, Junjun He, Tong Lu, Jifeng Dai, and Yu
  Qiao.
\newblock Vision transformer adapter for dense predictions.
\newblock In {\em ICLR}, 2023.

\bibitem{mask2former}
Bowen Cheng, Ishan Misra, Alexander~G Schwing, Alexander Kirillov, and Rohit
  Girdhar.
\newblock Masked-attention mask transformer for universal image segmentation.
\newblock In {\em CVPR}, pages 1290--1299, 2022.

\bibitem{deepglobe}
Ilke Demir, Krzysztof Koperski, David Lindenbaum, Guan Pang, Jing Huang, Saikat
  Basu, Forest Hughes, Devis Tuia, and Ramesh Raskar.
\newblock Deepglobe 2018: A challenge to parse the earth through satellite
  images.
\newblock In {\em CVPRW}, pages 172--181, 2018.

\bibitem{imagenet}
Jia Deng, Wei Dong, Richard Socher, Li-Jia Li, Kai Li, and Li Fei-Fei.
\newblock Imagenet: A large-scale hierarchical image database.
\newblock In {\em CVPR}, pages 248--255, 2009.

\bibitem{cell}
Ruining Deng, Can Cui, Quan Liu, Tianyuan Yao, Lucas~W Remedios, Shunxing Bao,
  Bennett~A Landman, Lee~E Wheless, Lori~A Coburn, Keith~T Wilson, et~al.
\newblock Segment anything model (sam) for digital pathology: Assess zero-shot
  segmentation on whole slide imaging.
\newblock {\em arXiv preprint arXiv:2304.04155}, 2023.

\bibitem{resunet}
Foivos~I Diakogiannis, Fran{\c{c}}ois Waldner, Peter Caccetta, and Chen Wu.
\newblock Resunet-a: A deep learning framework for semantic segmentation of
  remotely sensed data.
\newblock {\em ISPRS Journal of Photogrammetry and Remote Sensing},
  162:94--114, 2020.

\bibitem{dotav2}
Jian Ding, Nan Xue, Gui-Song Xia, Xiang Bai, Wen Yang, Michael Yang, Serge
  Belongie, Jiebo Luo, Mihai Datcu, Marcello Pelillo, and Liangpei Zhang.
\newblock Object detection in aerial images: A large-scale benchmark and
  challenges.
\newblock {\em IEEE Transactions on Pattern Analysis and Machine Intelligence},
  pages 1--1, 2021.

\bibitem{lanet}
Lei Ding, Hao Tang, and Lorenzo Bruzzone.
\newblock Lanet: Local attention embedding to improve the semantic segmentation
  of remote sensing images.
\newblock {\em IEEE Transactions on Geoscience and Remote Sensing},
  59(1):426--435, 2020.

\bibitem{vit}
Alexey Dosovitskiy, Lucas Beyer, Alexander Kolesnikov, Dirk Weissenborn,
  Xiaohua Zhai, Thomas Unterthiner, Mostafa Dehghani, Matthias Minderer, Georg
  Heigold, Sylvain Gelly, Jakob Uszkoreit, and Neil Houlsby.
\newblock An image is worth 16x16 words: Transformers for image recognition at
  scale.
\newblock {\em ICLR}, 2021.

\bibitem{mae}
Kaiming He, Xinlei Chen, Saining Xie, Yanghao Li, Piotr Doll\'ar, and Ross
  Girshick.
\newblock Masked autoencoders are scalable vision learners.
\newblock In {\em CVPR}, pages 16000--16009, June 2022.

\bibitem{resnet}
Kaiming He, Xiangyu Zhang, Shaoqing Ren, and Jian Sun.
\newblock Deep residual learning for image recognition.
\newblock In {\em CVPR}, pages 770--778, 2016.

\bibitem{stunet}
Xin He, Yong Zhou, Jiaqi Zhao, Di Zhang, Rui Yao, and Yong Xue.
\newblock Swin transformer embedding unet for remote sensing image semantic
  segmentation.
\newblock {\em IEEE Transactions on Geoscience and Remote Sensing}, 60:1--15,
  2022.

\bibitem{sam_mars}
Sahib Julka and Michael Granitzer.
\newblock Knowledge distillation with segment anything (sam) model for
  planetary geological mapping.
\newblock {\em arXiv preprint arXiv:2305.07586}, 2023.

\bibitem{sam}
Alexander Kirillov, Eric Mintun, Nikhila Ravi, Hanzi Mao, Chloe Rolland, Laura
  Gustafson, Tete Xiao, Spencer Whitehead, Alexander~C. Berg, Wan-Yen Lo, Piotr
  Dollar, and Ross Girshick.
\newblock Segment anything.
\newblock In {\em ICCV}, pages 4015--4026, October 2023.

\bibitem{dior}
Ke Li, Gang Wan, Gong Cheng, Liqiu Meng, and Junwei Han.
\newblock Object detection in optical remote sensing images: A survey and a new
  benchmark.
\newblock {\em ISPRS journal of photogrammetry and remote sensing},
  159:296--307, 2020.

\bibitem{coco}
Tsung-Yi Lin, Michael Maire, Serge Belongie, James Hays, Pietro Perona, Deva
  Ramanan, Piotr Doll{\'a}r, and C~Lawrence Zitnick.
\newblock Microsoft coco: Common objects in context.
\newblock In {\em ECCV}, pages 740--755, 2014.

\bibitem{groundingdino}
Shilong Liu, Zhaoyang Zeng, Tianhe Ren, Feng Li, Hao Zhang, Jie Yang, Chunyuan
  Li, Jianwei Yang, Hang Su, Jun Zhu, et~al.
\newblock Grounding dino: Marrying dino with grounded pre-training for open-set
  object detection.
\newblock {\em arXiv preprint arXiv:2303.05499}, 2023.

\bibitem{swint}
Ze Liu, Yutong Lin, Yue Cao, Han Hu, Yixuan Wei, Zheng Zhang, Stephen Lin, and
  Baining Guo.
\newblock {Swin Transformer}: Hierarchical vision transformer using shifted
  windows.
\newblock In {\em ICCV}, pages 10012--10022, 2021.

\bibitem{hrsc2016}
Zikun Liu, Liu Yuan, Lubin Weng, and Yiping Yang.
\newblock A high resolution optical satellite image dataset for ship
  recognition and some new baselines.
\newblock In {\em ICPRAM}, pages 324--331, 2017.

\bibitem{millionaid}
Yang Long, Gui-Song Xia, Shengyang Li, Wen Yang, Michael~Ying Yang, Xiao~Xiang
  Zhu, Liangpei Zhang, and Deren Li.
\newblock On creating benchmark dataset for aerial image interpretation:
  Reviews, guidances, and million-aid.
\newblock {\em IEEE Journal of Selected Topics in Applied Earth Observations
  and Remote Sensing}, 14:4205--4230, 2021.

\bibitem{uavid}
Ye Lyu, George Vosselman, Gui-Song Xia, Alper Yilmaz, and Michael~Ying Yang.
\newblock Uavid: A semantic segmentation dataset for uav imagery.
\newblock {\em ISPRS journal of photogrammetry and remote sensing},
  165:108--119, 2020.

\bibitem{ass_2022_tgrs_factseg}
Ailong Ma, Junjue Wang, Yanfei Zhong, and Zhuo Zheng.
\newblock Factseg: Foreground activation-driven small object semantic
  segmentation in large-scale remote sensing imagery.
\newblock {\em IEEE Transactions on Geoscience and Remote Sensing}, 60:1--16,
  2022.

\bibitem{medical}
Jun Ma and Bo Wang.
\newblock Segment anything in medical images.
\newblock {\em arXiv preprint arXiv:2304.12306}, 2023.

\bibitem{zeebruge}
Diego Marcos, Michele Volpi, Benjamin Kellenberger, and Devis Tuia.
\newblock Land cover mapping at very high resolution with rotation equivariant
  cnns: Towards small yet accurate models.
\newblock {\em ISPRS journal of photogrammetry and remote sensing},
  145:96--107, 2018.

\bibitem{hmanet}
Ruigang Niu, Xian Sun, Yu Tian, Wenhui Diao, Kaiqiang Chen, and Kun Fu.
\newblock Hybrid multiple attention network for semantic segmentation in aerial
  images.
\newblock {\em IEEE Transactions on Geoscience and Remote Sensing}, 60:1--18,
  2022.

\bibitem{sam_oneshot}
Lucas~Prado Osco, Qiusheng Wu, Eduardo~Lopes de Lemos, Wesley~Nunes
  Gon{\c{c}}alves, Ana Paula~Marques Ramos, Jonathan Li, and Jos{\'e}~Marcato
  Junior.
\newblock The segment anything model (sam) for remote sensing applications:
  From zero to one shot.
\newblock {\em arXiv preprint arXiv:2306.16623}, 2023.

\bibitem{clip}
Alec Radford, Jong~Wook Kim, Chris Hallacy, Aditya Ramesh, Gabriel Goh,
  Sandhini Agarwal, Girish Sastry, Amanda Askell, Pamela Mishkin, Jack Clark,
  et~al.
\newblock Learning transferable visual models from natural language
  supervision.
\newblock In {\em ICML}, pages 8748--8763. PMLR, 2021.

\bibitem{sam_space}
Simiao Ren, Francesco Luzi, Saad Lahrichi, Kaleb Kassaw, Leslie~M Collins, Kyle
  Bradbury, and Jordan~M Malof.
\newblock Segment anything, from space?
\newblock {\em arXiv preprint arXiv:2304.13000}, 2023.

\bibitem{unet}
Olaf Ronneberger, Philipp Fischer, and Thomas Brox.
\newblock U-net: Convolutional networks for biomedical image segmentation.
\newblock In {\em MICCAI}, pages 234--241, 2015.

\bibitem{fair1m}
Xian Sun, Peijin Wang, Zhiyuan Yan, Feng Xu, Ruiping Wang, Wenhui Diao, Jin
  Chen, Jihao Li, Yingchao Feng, Tao Xu, Martin Weinmann, Stefan Hinz, Cheng
  Wang, and Kun Fu.
\newblock Fair1m: A benchmark dataset for fine-grained object recognition in
  high-resolution remote sensing imagery.
\newblock {\em ISPRS Journal of Photogrammetry and Remote Sensing},
  184:116--130, 2022.

\bibitem{gid}
Xin-Yi Tong, Gui-Song Xia, Qikai Lu, Huanfeng Shen, Shengyang Li, Shucheng You,
  and Liangpei Zhang.
\newblock Land-cover classification with high-resolution remote sensing images
  using transferable deep models.
\newblock {\em Remote Sensing of Environment}, 237:111322, 2020.

\bibitem{zurich_summer}
Michele Volpi and Vittorio Ferrari.
\newblock Semantic segmentation of urban scenes by learning local class
  interactions.
\newblock In {\em CVPRW}, pages 1--9, 2015.

\bibitem{rsp}
Di Wang, Jing Zhang, Bo Du, Gui-Song Xia, and Dacheng Tao.
\newblock An empirical study of remote sensing pretraining.
\newblock {\em IEEE Transactions on Geoscience and Remote Sensing}, 61:1--20,
  2023.

\bibitem{rvsa}
Di Wang, Qiming Zhang, Yufei Xu, Jing Zhang, Bo Du, Dacheng Tao, and Liangpei
  Zhang.
\newblock Advancing plain vision transformer toward remote sensing foundation
  model.
\newblock {\em IEEE Transactions on Geoscience and Remote Sensing}, 61:1--15,
  2023.

\bibitem{loveda}
Junjue Wang, Zhuo Zheng, Ailong Ma, Xiaoyan Lu, and Yanfei Zhong.
\newblock Loveda: A remote sensing land-cover dataset for domain adaptive
  semantic segmentation.
\newblock In {\em NeurIPS Track on Datasets and Benchmarks}, volume~1, 2021.

\bibitem{dc_swin}
Libo Wang, Rui Li, Chenxi Duan, Ce Zhang, Xiaoliang Meng, and Shenghui Fang.
\newblock A novel transformer based semantic segmentation scheme for
  fine-resolution remote sensing images.
\newblock {\em IEEE Geoscience and Remote Sensing Letters}, 19:1--5, 2022.

\bibitem{internimage}
Wenhai Wang, Jifeng Dai, Zhe Chen, Zhenhang Huang, Zhiqi Li, Xizhou Zhu,
  Xiaowei Hu, Tong Lu, Lewei Lu, Hongsheng Li, et~al.
\newblock Internimage: Exploring large-scale vision foundation models with
  deformable convolutions.
\newblock In {\em CVPR}, pages 14408--14419, 2023.

\bibitem{isaid}
Syed Waqas~Zamir, Aditya Arora, Akshita Gupta, Salman Khan, Guolei Sun, Fahad
  Shahbaz~Khan, Fan Zhu, Ling Shao, Gui-Song Xia, and Xiang Bai.
\newblock isaid: A large-scale dataset for instance segmentation in aerial
  images.
\newblock In {\em CVPRW}, pages 28--37, 2019.

\bibitem{dota}
Gui-Song Xia, Xiang Bai, Jian Ding, Zhen Zhu, Serge Belongie, Jiebo Luo, Mihai
  Datcu, Marcello Pelillo, and Liangpei Zhang.
\newblock Dota: A large-scale dataset for object detection in aerial images.
\newblock In {\em CVPR}, June 2018.

\bibitem{upernet}
Tete Xiao, Yingcheng Liu, Bolei Zhou, Yuning Jiang, and Jian Sun.
\newblock Unified perceptual parsing for scene understanding.
\newblock In {\em ECCV}, pages 418--434, 2018.

\bibitem{rssformer}
Rongtao Xu, Changwei Wang, Jiguang Zhang, Shibiao Xu, Weiliang Meng, and
  Xiaopeng Zhang.
\newblock Rssformer: Foreground saliency enhancement for remote sensing
  land-cover segmentation.
\newblock {\em IEEE Transactions on Image Processing}, 32:1052--1064, 2023.

\bibitem{sam_text2seg}
Jielu Zhang, Zhongliang Zhou, Gengchen Mai, Lan Mu, Mengxuan Hu, and Sheng Li.
\newblock Text2seg: Remote sensing image semantic segmentation via text-guided
  visual foundation models.
\newblock {\em arXiv preprint arXiv:2304.10597}, 2023.

\bibitem{vitae_v2}
Qiming Zhang, Yufei Xu, Jing Zhang, and Dacheng Tao.
\newblock Vitaev2: Vision transformer advanced by exploring inductive bias for
  image recognition and beyond.
\newblock {\em International Journal of Computer Vision}, 131(5):1141--1162,
  2023.

\bibitem{persam}
Renrui Zhang, Zhengkai Jiang, Ziyu Guo, Shilin Yan, Junting Pan, Hao Dong, Peng
  Gao, and Hongsheng Li.
\newblock Personalize segment anything model with one shot.
\newblock {\em arXiv preprint arXiv:2305.03048}, 2023.

\bibitem{eanet}
Xianwei Zheng, Linxi Huan, Gui-Song Xia, and Jianya Gong.
\newblock Parsing very high resolution urban scene images by learning deep
  convnets with edge-aware loss.
\newblock {\em ISPRS Journal of Photogrammetry and Remote Sensing}, 170:15--28,
  2020.

\end{thebibliography}

\newpage

\appendix

\section{Appendix}

\subsection{Category Abbreviations}

For the SOTA dataset, we present the list of all category abbreviations as follows. \textit{LV: large vehicle, SP: swimming pool, HC: helicopter, BR: bridge, PL: plane, SH: ship, SBF: soccer ball field, BC: basketball court, GTF: ground track field, SV: small vehicle, BD: baseball diamond, TC: tennis court, RA: roundabout, ST: storage tank, HA: harbor, CC: container crane, AP: airport, HP: helipad.}

For the SIOR dataset, we present the list of all category abbreviations as follows. \textit{APL: airplane, APO: airport, BF: baseball field, BC: basketball court, BR: bridge, CH: chimney, ESA: expressway service area, ETS: expressway toll station, DA: dam, GF: golf field, GTF: ground track field, HA: harbor, OP: overpass, SH: ship, STD: stadium, STT: storage tank, TC: tennis court, TS: train station, VH: vehicle, WD: windmill.}

For the FAST dataset, we present the list of all category abbreviations as follows. \textit{A2: A220, A3: A321, A4: A330, A5: A350, ARJ: ARJ21, BF: baseball field, BC: basketball court, B3: boeing737, B4: boeing747, B7: boeing777, B8: boeing787, BR: bridge, BU: bus, C9: C919, CT: cargo truck, DCS: dry cargo ship, DT: dump truck, ES: engineering ship, EV: excavator, FB: fishing boat, FF: football field', IN: intersection, LCS: liquid cargo ship, MB: motorboat, OA: other airplane, OS: other ship, OV: other vehicle, PS: passenger ship, RA: roundabout, SC: small car, TC: tennis court, TRT: tractor, TRL: trailer, TUT: truck tractor, TB: tugboat, VA: van, WS: warship.}

\subsection{Experiment Settings}

We present the experiment settings of pre-training and fine-tuning in Table~\ref{optim_1}-\ref{optim_2}.

\begin{table}[h]
  \small
  \begin{minipage}{0.4\linewidth}
  \caption{Basic settings in experiments. "|" means pre-training | fine-tuning.}
  \centering
    \begin{tabular}{l|c}
  \hline
   Config & value \\
    \hline
   Optimizer & AdamW  \\
   Momentum  & (0.9, 0.999)\\
   Batchsize & 96 | 8 \\
   Iterations & 80000 \\
   Scheduler & cosine decay \\
   \hline

\end{tabular}
\label{optim_1}
\end{minipage}
\hspace{2mm}
\begin{minipage}{0.6\linewidth}
  \caption{Detailed settings of different models. "|" means pre-training | fine-tuning. ILR: Initial learning rate. WD: Weight decay. MLR: Minimum learning rate}
  \centering
    \begin{tabular}{l|ccc}
  \hline
   Backbone & ILR  & WD & MLR \\
    \hline
   ResNet \cite{resnet} & 1e-3  & 5e-2 | 1e-4  & 5e-6 \\
   Swin \cite{swint} /ViTAEv2 \cite{vitae_v2} & 6e-5 & 1e-2 & 0 \\
   InternImage \cite{internimage} & 6e-5 & 5e-2 & 0 \\
   ViT \cite{vit} /ViT-RVSA \cite{rvsa} & 6e-5 & 5e-2 & 0 \\
   ViT-Adapter \cite{vit_adapter} & 6e-5 & 1e-2 & 0 \\
   \hline

\end{tabular}
\label{optim_2}
\end{minipage}
\end{table}

\subsection{SAMRS Training and Validation sets}

For the experiments based on the SAMRS dataset (see Table 2 in the main text), in each subset, 95\% samples are used for pre-training. They together consist of the SAMRS training set. The remained samples consist of the SAMRS validation set. SAMRS training and validation sets have 88,685 and 4,667 images, respectively. The samples transformed from the DIOR testing set \cite{dior} have not been used in any experiment.

\subsection{SAMRS-MAE}

We conduct MAE pre-training \cite{mae} on the SAMRS training set. To improve the pre-training performance, we further clip the image to 384 $\times$ 384 with a stride of 300, obtaining 609,707 images.

\subsection{Evaluations on the SAMRS Validation Set}

Table~\ref{val} lists the evaluation results of the pre-trained models on the SAMRS validation sets. For convenience, We uniformly use the images in the size of 512 $\times$ 512 obtained through a center cropping on SAMRS validation set samples for evaluation. Here, mIOU is adopted as the metric. It can be seen that all scores on the FAST validation set are relatively low, indicating the challenging nature of the proposed dataset. By comparing Table \ref{val} to the fine-tuning results (Table 3-4 in the main text), we can find that model performance on validation and fine-tuning show similar trends. For example, the performances of adopting SEP alone are not as well as with a good initialation. As can be seen, compared to UNet, UperNet achieves higher accuracies, indicating the model representations are more expressive. Therefore, UperNet can obtain better performances on the challenging IsAID dataset. In addition, it can be observed that the performances of vision transformer networks still surpass convolutional networks, especially for the hierarchical structure. We notice the InternImage performs poorly with 512 $\times$ 512 images. Therefore, we resize the input image to 224 $\times$ 224, and the accuracy is recovered. Note the setting of 224 $\times$ 224 is adopted in pre-training. These results indicate the InternImage may be more dependent on the input size of the pre-training. We have not presented the results of Mask2Former \cite{mask2former} because the validation accuracies on three subsets are close to 0, implying serious over fittings in pre-training since the loss was continuously decreasing. Compared to UNet and UperNet, Mask2former is a newly-proposed framework, it has many different hyperparameters that need to be carefully tuned. Further investigation of hyperparameter settings is required to evaluate its impact on remote sensing images. Nevertheless, we believe the SAMRS validation set is expected to help adjust the settings of SEP in future explorations.

\begin{table}[h]
  \small
  \caption{The mIoUs of different models on SAMRS validation set. $\dagger$: The input image is resized to 224 $\times$ 224. $\dagger\dagger$: MAE pre-training on the MillionAID. $\dagger\dagger\dagger$: MAE pre-training on the SAMRS training set}
  \newcommand{\tabincell}[2]{\begin{tabular}{@{}#1@{}}#2\end{tabular}}
  \centering
    \begin{tabular}{lll|cccc}
  \hline
    Method & Pretrain & Backbone   & SOTA & SIOR & FAST & Average \\
    \hline

    UNet \cite{unet} & SEP     & ResNet-50  & 55.88 & 71.44 & 35.35 & 54.23 \\
    UNet & IMP \cite{imagenet}+SEP & ResNet-50  & 58.62 & 80.49 & 37.20 & 58.77 \\
    UNet & RSP \cite{rsp}+SEP & ResNet-50  & 62.43 & 85.37 & 38.41 & 62.07 \\
    \hline
    UperNet \cite{upernet} & SEP     & ResNet-50  & 64.79 & 82.74 & 39.32 & 62.28 \\
    UperNet & IMP+SEP & ResNet-50  & 73.59 & 88.59 & 47.10 & 69.76  \\
    UperNet & RSP+SEP & ResNet-50  & 76.03 & 90.66 & 47.62 & 71.44 \\
    \hline
    UperNet & IMP+SEP & Swin-T & 81.53 & 94.64 & 57.91 & 78.03 \\
    UperNet & IMP+SEP & ViTAEv2-S & 81.13 & 94.06 & 57.90 & 77.70 \\
    UperNet & IMP+SEP & InternImage-T & 58.07 & 72.04 & 29.33 & 53.15 \\
    UperNet & IMP+SEP & InternImage-T $\dagger$ & 78.06 & 92.51 & 45.70 & 72.09 \\
    \hline
    UperNet & IMP+SEP & ViT-B & 79.37 & 92.76 & 53.05 & 75.06 \\
    UperNet & IMP+SEP & ViT-Adapter-B & 80.41 & 93.76 & 51.13 & 75.10 \\
    UperNet & BEIT \cite{beit} +SEP & ViT-B & 74.10 & 85.81 & 42.04 & 67.31 \\
    UperNet & MAE \cite{mae} +SEP $\dagger\dagger$  & ViT-B + RVSA & 79.00 & 92.09 & 53.46 & 74.85 \\
    UperNet & SAMRS-MAE+SEP $\dagger\dagger\dagger$ & ViT-B + RVSA  & 77.64 & 91.87 & 54.78 & 74.67 \\
   \hline

\end{tabular}
\label{val}
\end{table}

\subsection{Dataset of Fine-tuning}

We fine-tune the pre-trained model on two commonly used RS segmentation datasets, including ISPRS Potsdam \footnote[1]{https://www.isprs.org/education/benchmarks/UrbanSemLab/2d-sem-label-potsdam.aspx} and iSAID \cite{isaid}. Before using them, we conduct a series of pre-processing. Here are the details.

\noindent\textbf{ISPRS Potsdam}: This is the most classical high-resolution RS segmentation dataset. It has 38 large images with an average size of 6,000 $\times$ 6,000, where the training and testing sets separately include 24 and 14 images. It contains 6 categories: impervious surface, building, low vegetation, tree, car, and clutter. In experiments, we crop the image into 512 $\times$ 512 with a stride of 320, obtaining 8,664 and 5,054 training and testing images. We only use RGB channels. Following most literature in the RS field \cite{rsp, eanet, lanet}, we ignore the clutter category in training and testing.

\noindent\textbf{iSAID}: This is a challenging dataset. It provides 15 foregrounds and 1 background category, where 2,806 high-resolution images that range from 800 $\times$ 800 to 4,000 $\times$ 13,000 pixels are contained. The training, validation, and test sets separately have 1,411, 458, and 937 large images. In this paper, we use the validation set for evaluation since the testing set cannot be acquired. In experiments, we crop the image into 896 $\times$ 896 by stride 512, increasing the size of the training and validation set to 33,620 and 11,533. In addition, only the foreground categories are considered.

\subsection{Visualization}

\begin{figure}[h]
    \centering
    \includegraphics[width=0.32\linewidth]{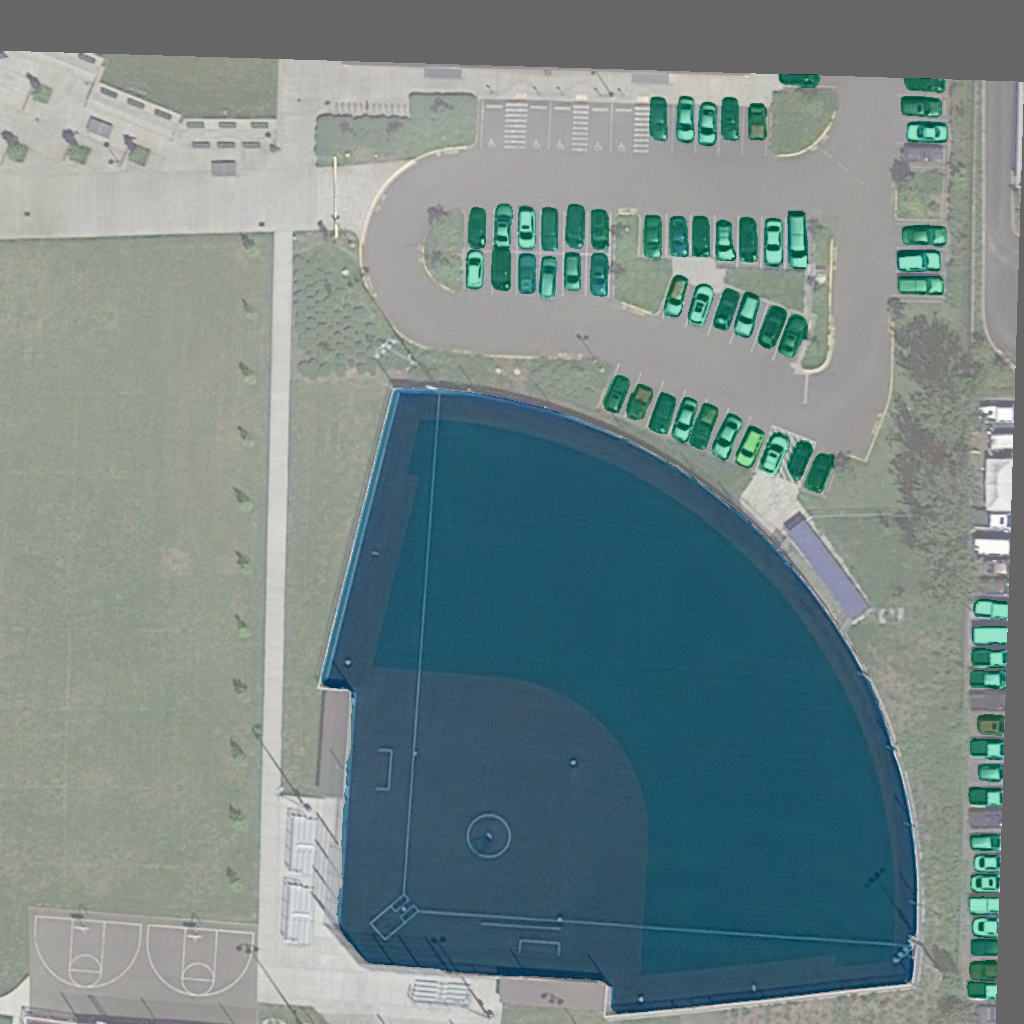}
    \includegraphics[width=0.32\linewidth]{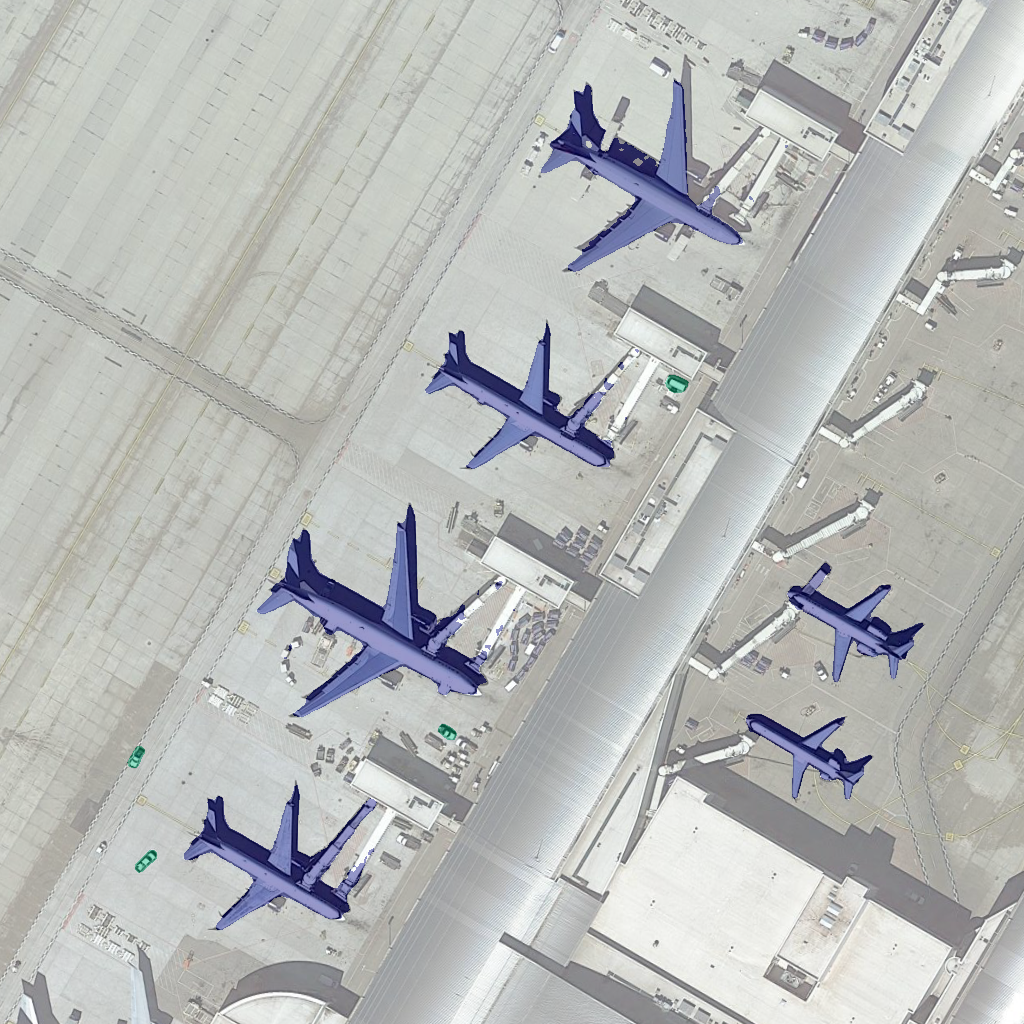}
    \includegraphics[width=0.32\linewidth]{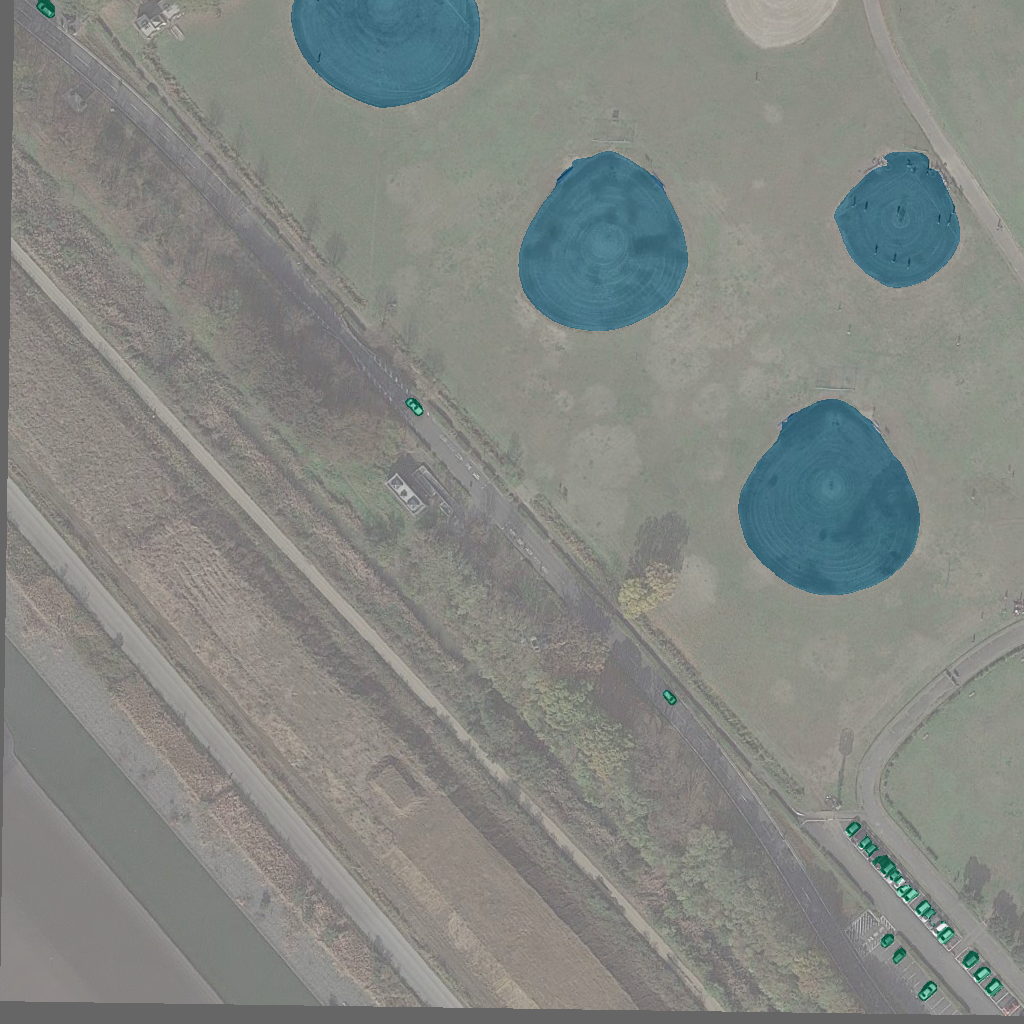}\\
    \includegraphics[width=0.32\linewidth]{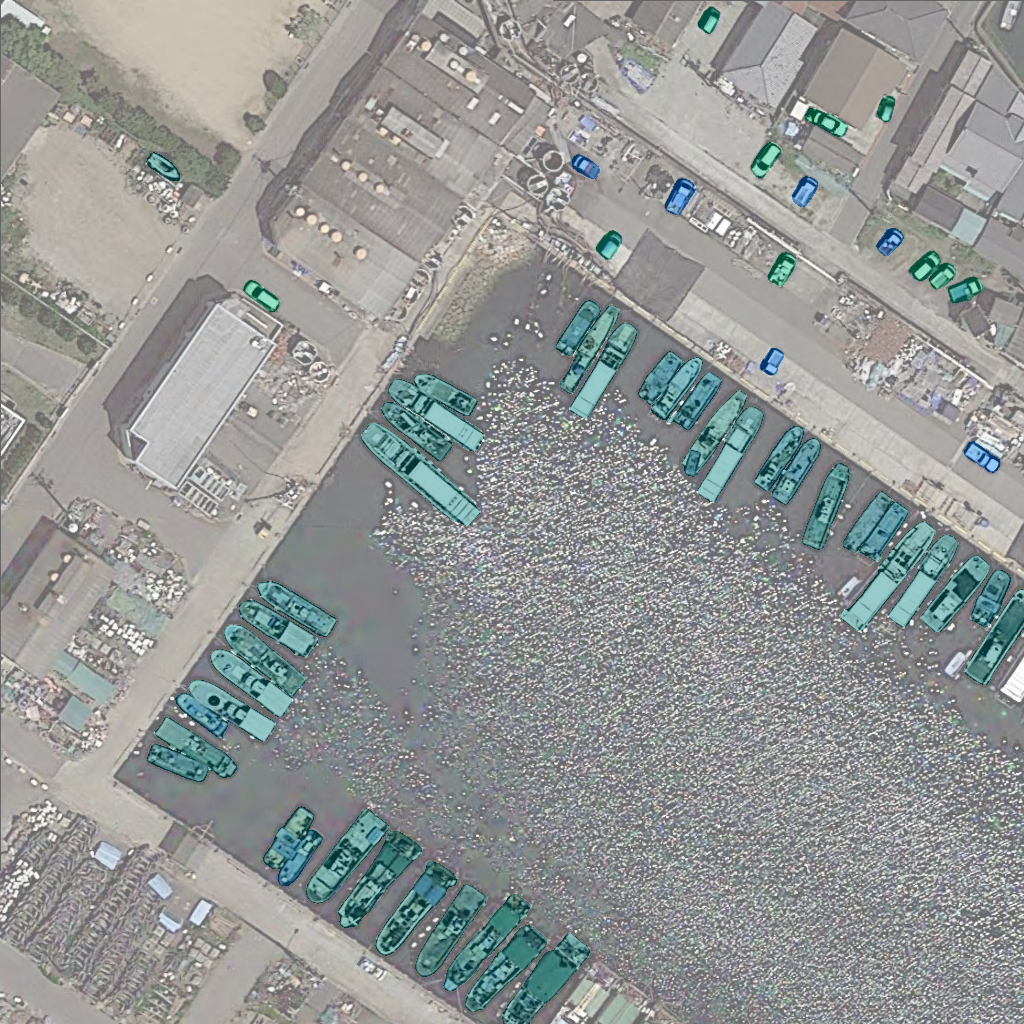}
    \includegraphics[width=0.32\linewidth]{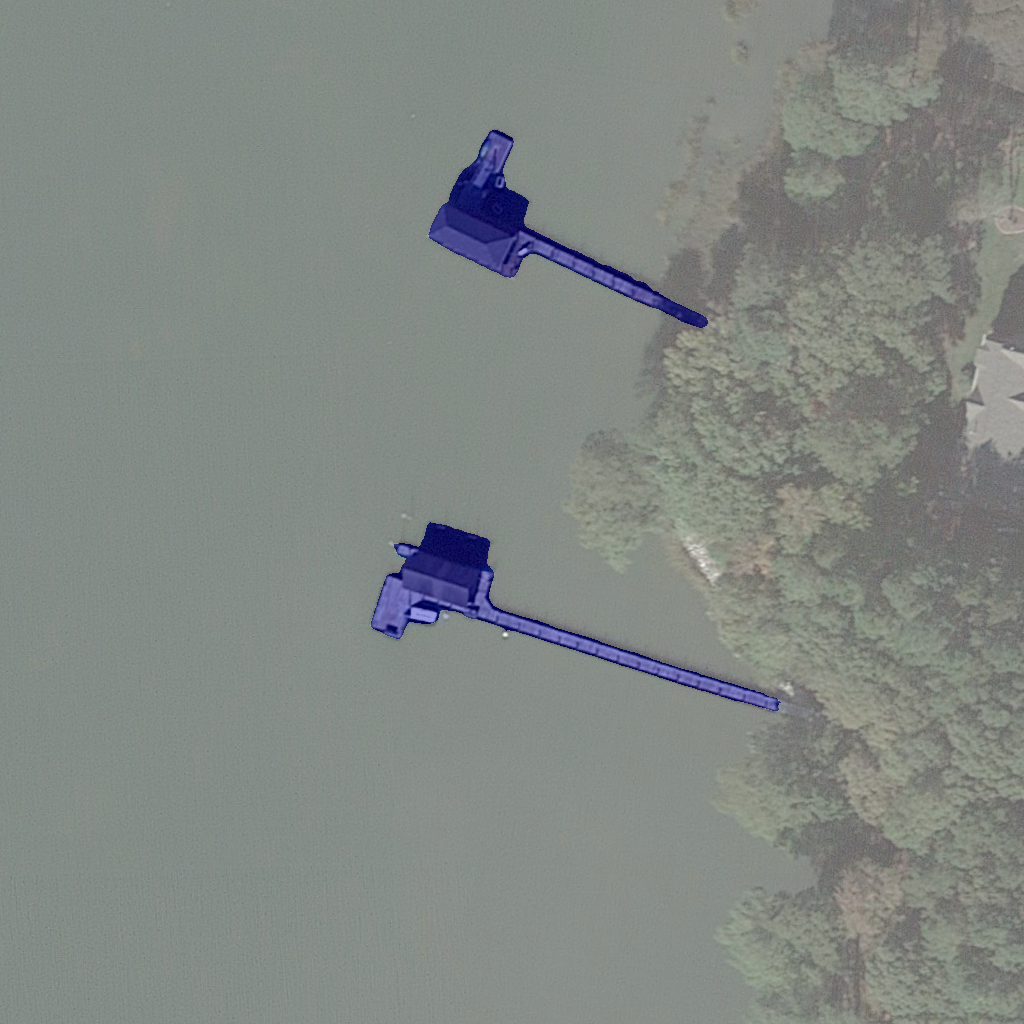}
    \includegraphics[width=0.32\linewidth]{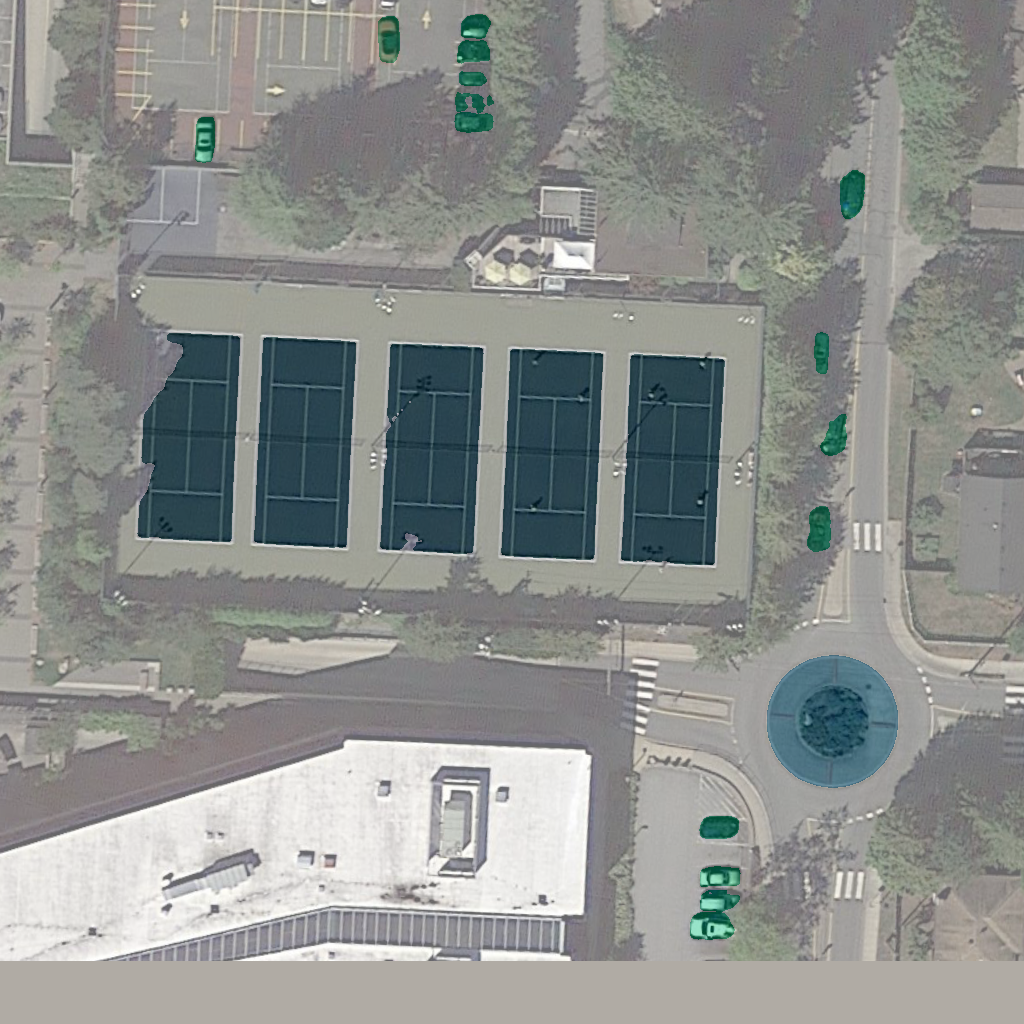}\\
    \includegraphics[width=0.32\linewidth]{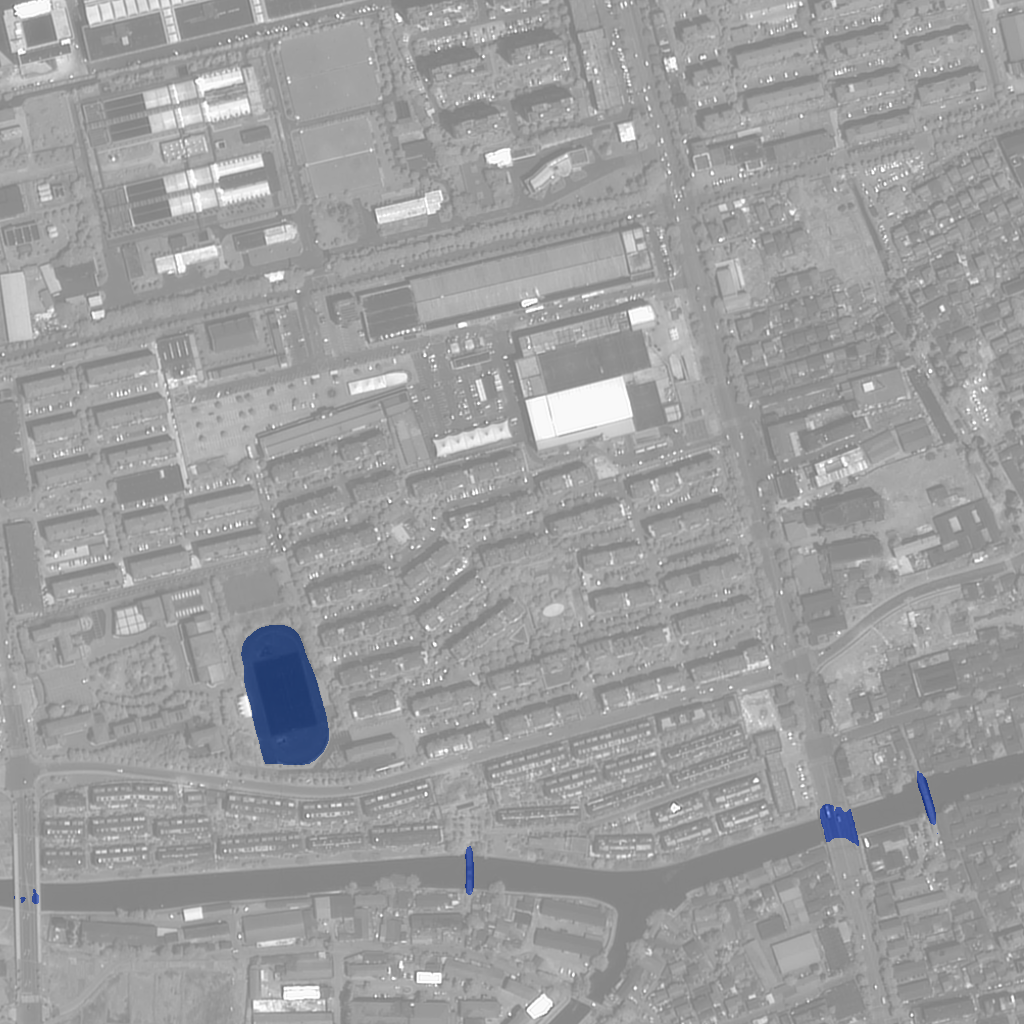}
    \includegraphics[width=0.32\linewidth]{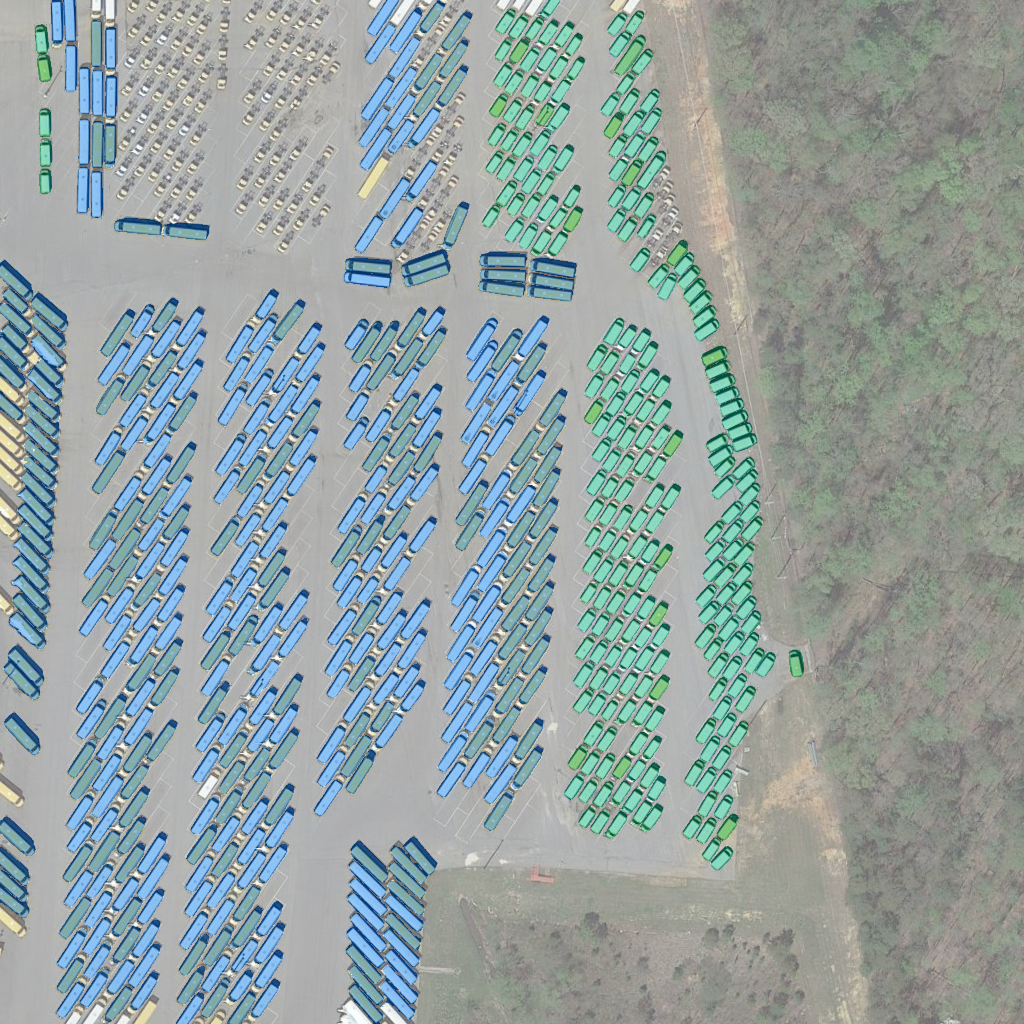}
    \includegraphics[width=0.32\linewidth]{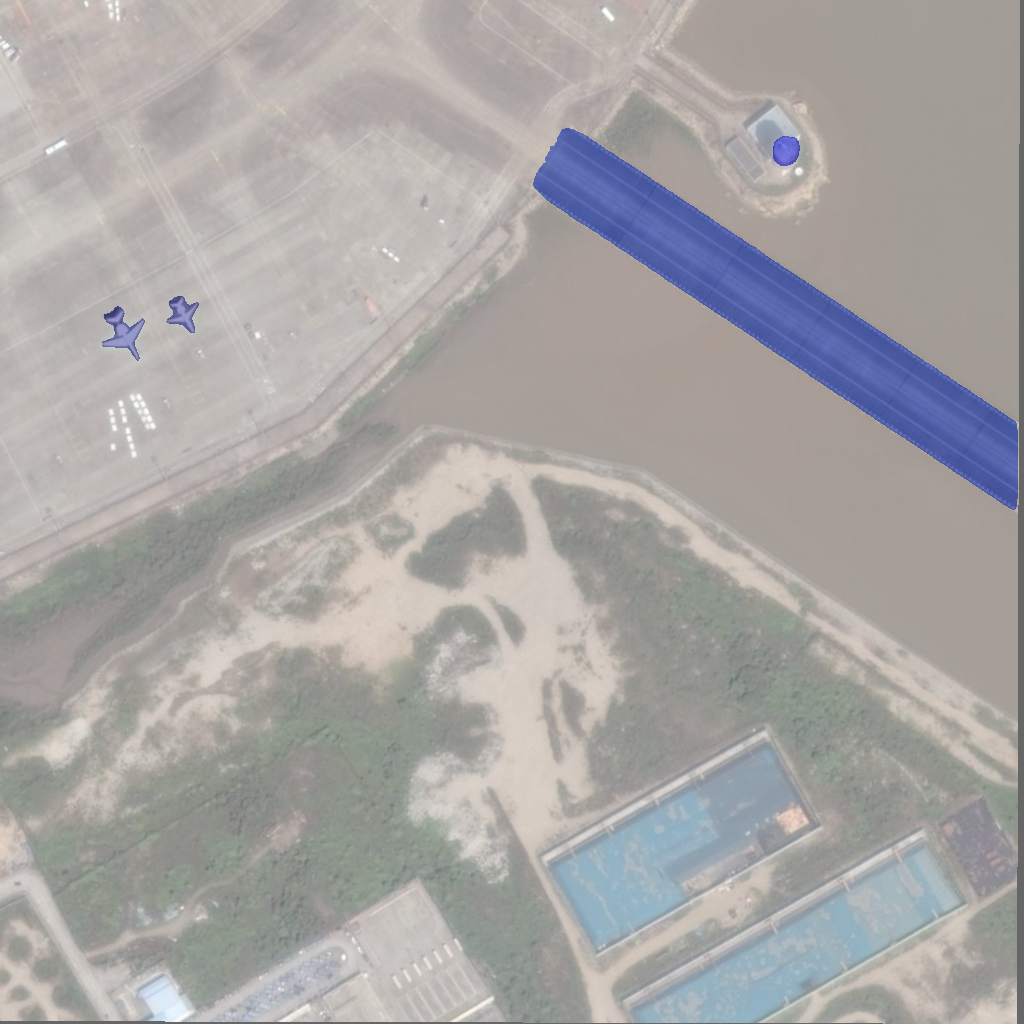}\\
    \includegraphics[width=0.32\linewidth]{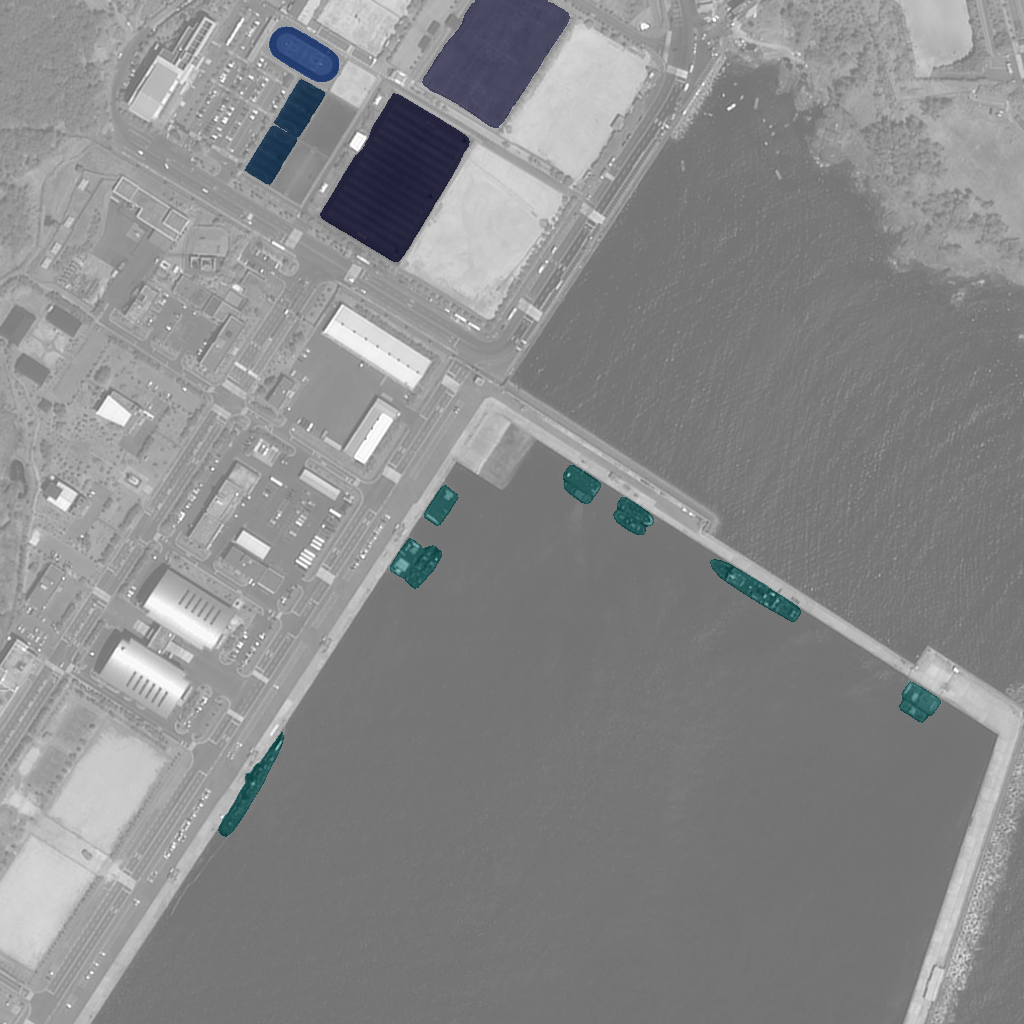}
    \includegraphics[width=0.32\linewidth]{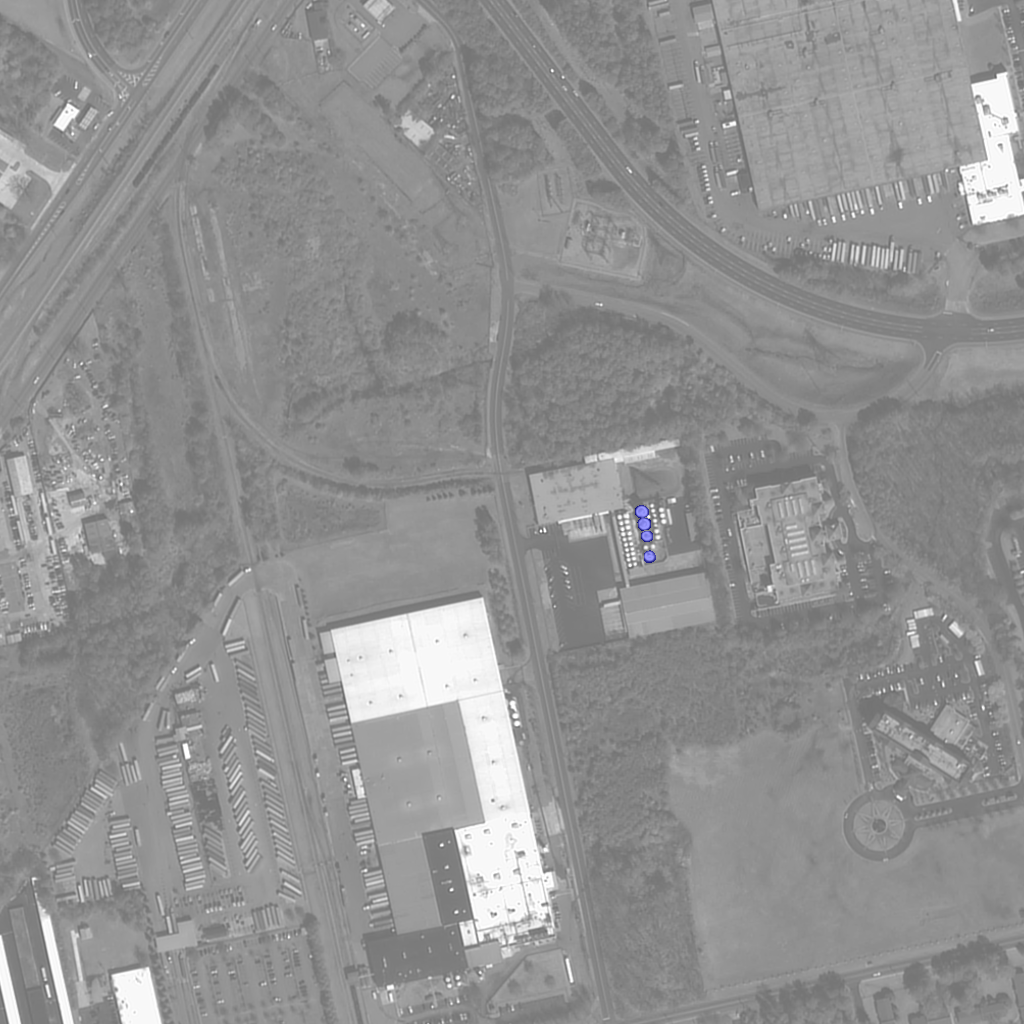}
    \includegraphics[width=0.32\linewidth]{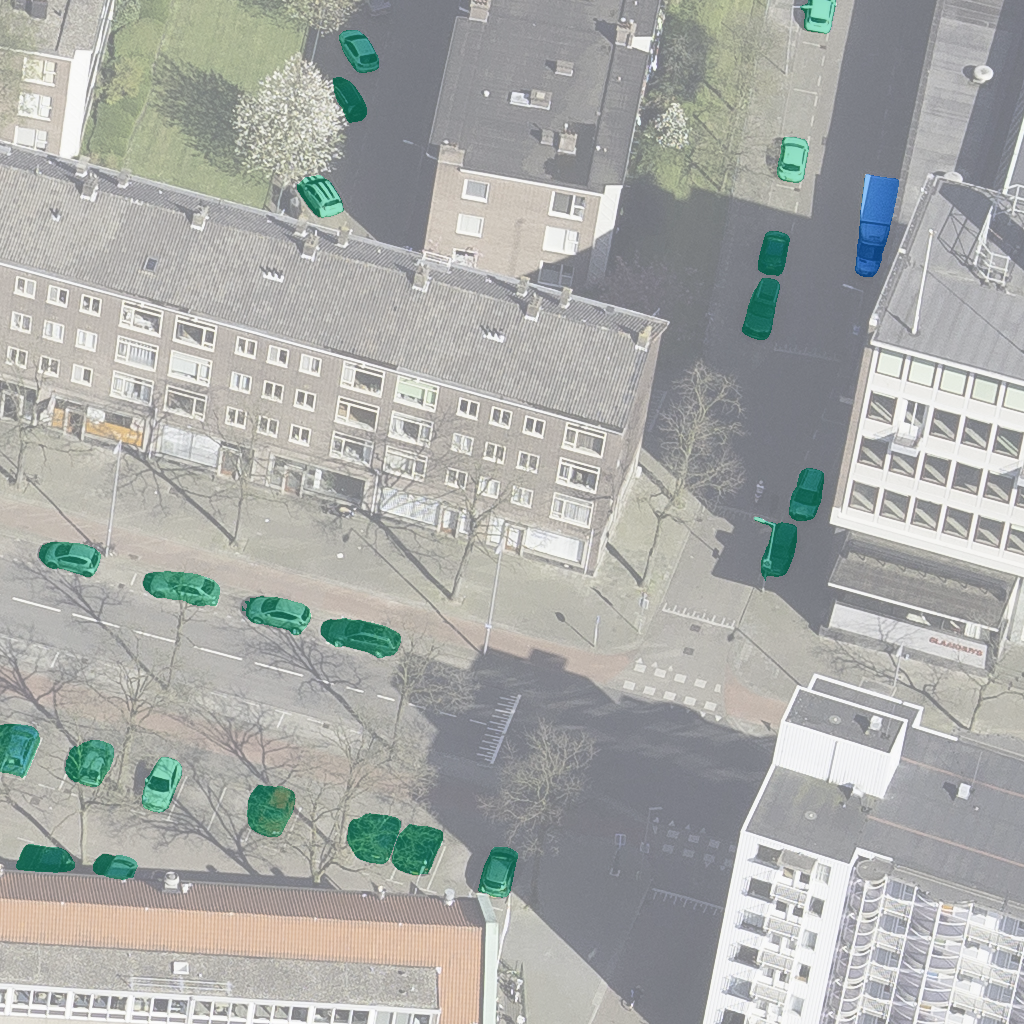}\\
    \caption{Visual examples from the SOTA subset of our SAMRS dataset.}
    \label{vis_sota}
\end{figure}

\begin{figure}[h]
    \centering
    \includegraphics[width=0.32\linewidth]{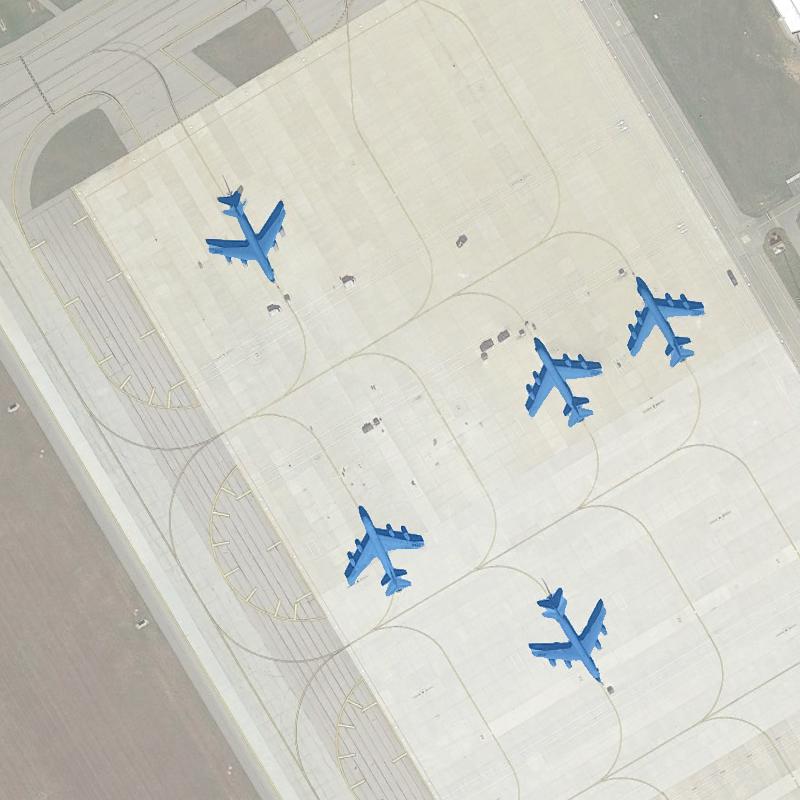}
    \includegraphics[width=0.32\linewidth]{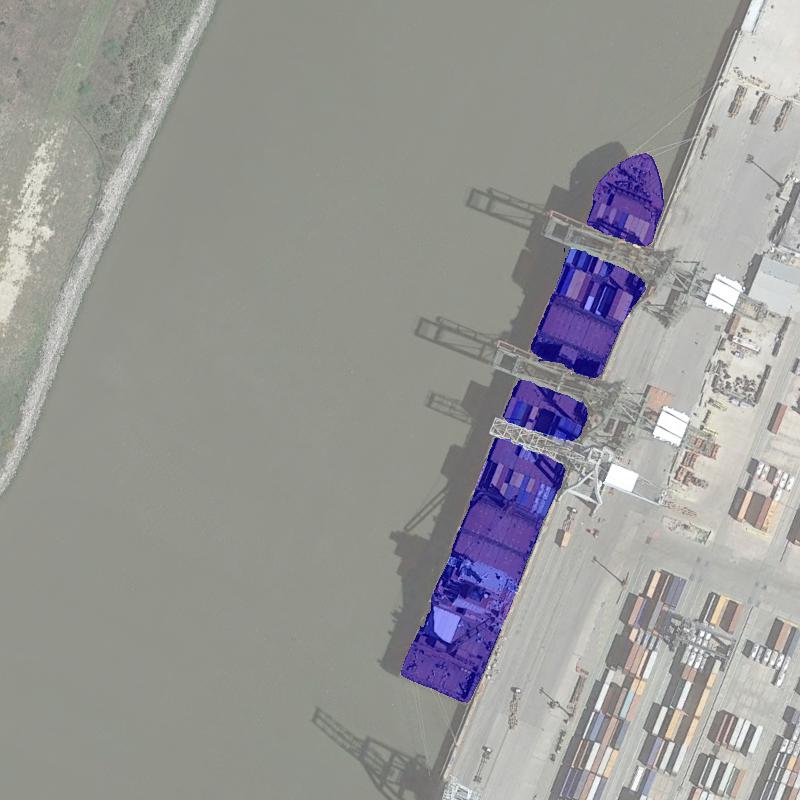}
    \includegraphics[width=0.32\linewidth]{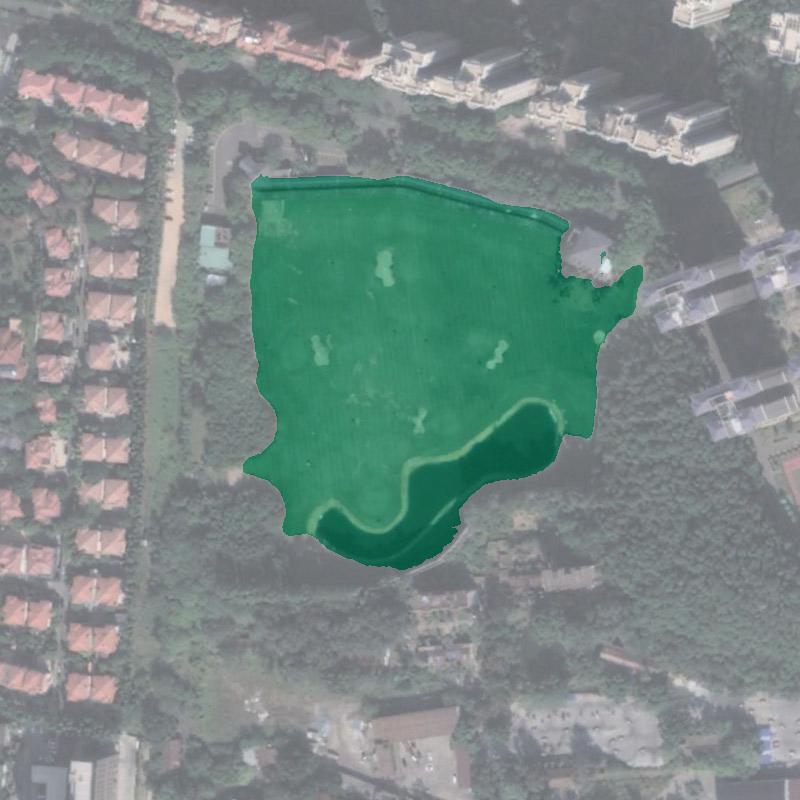}\\
    \includegraphics[width=0.32\linewidth]{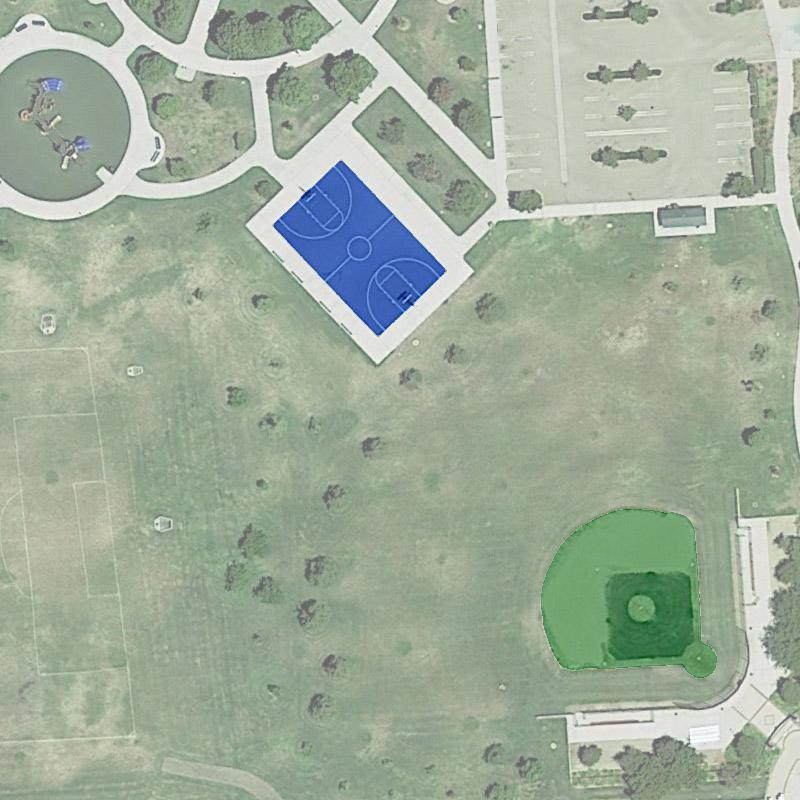}
    \includegraphics[width=0.32\linewidth]{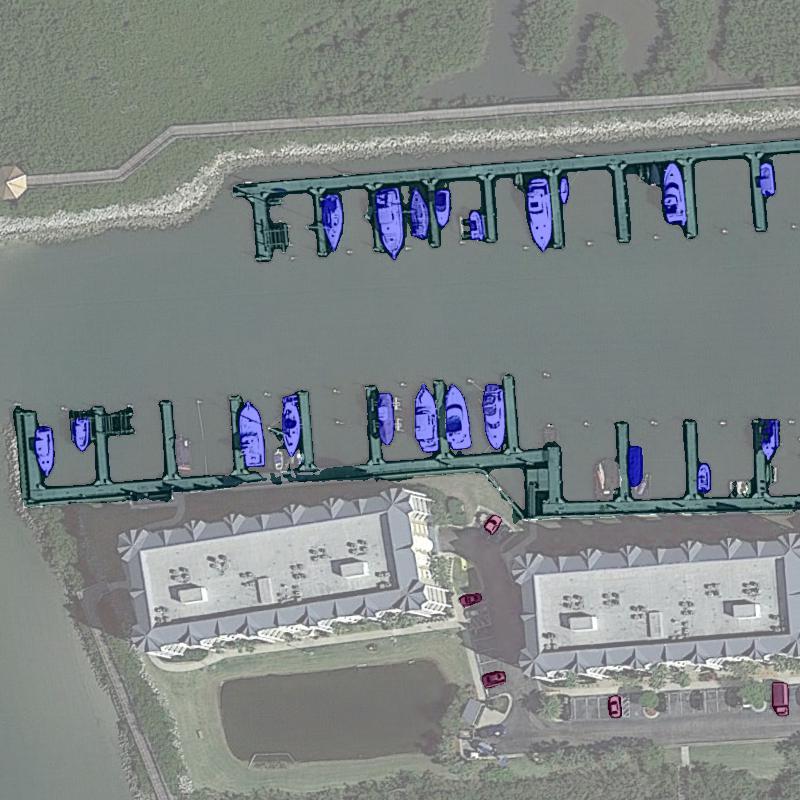}
    \includegraphics[width=0.32\linewidth]{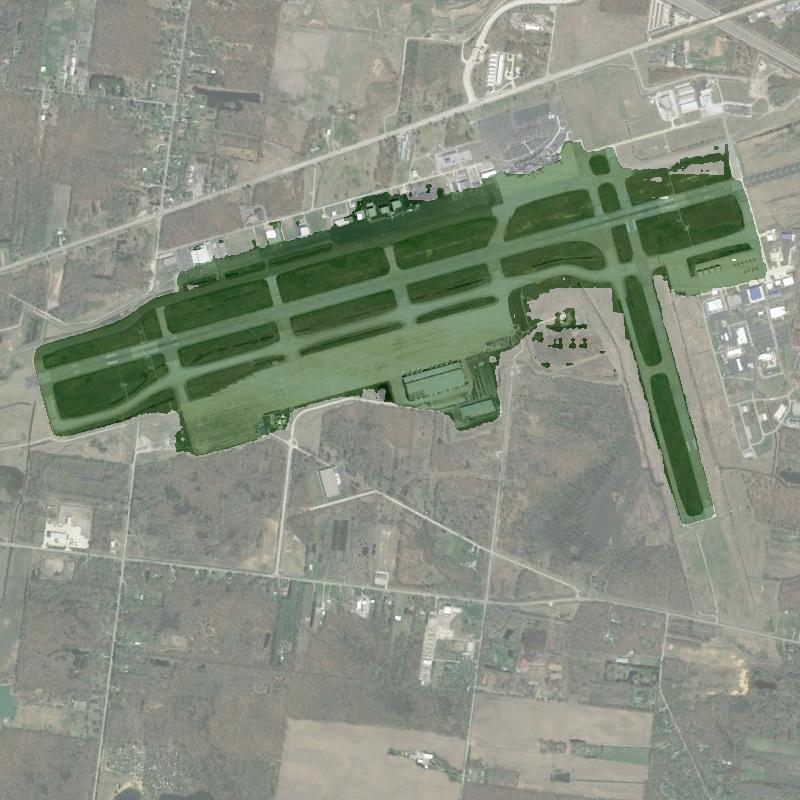}\\
    \includegraphics[width=0.32\linewidth]{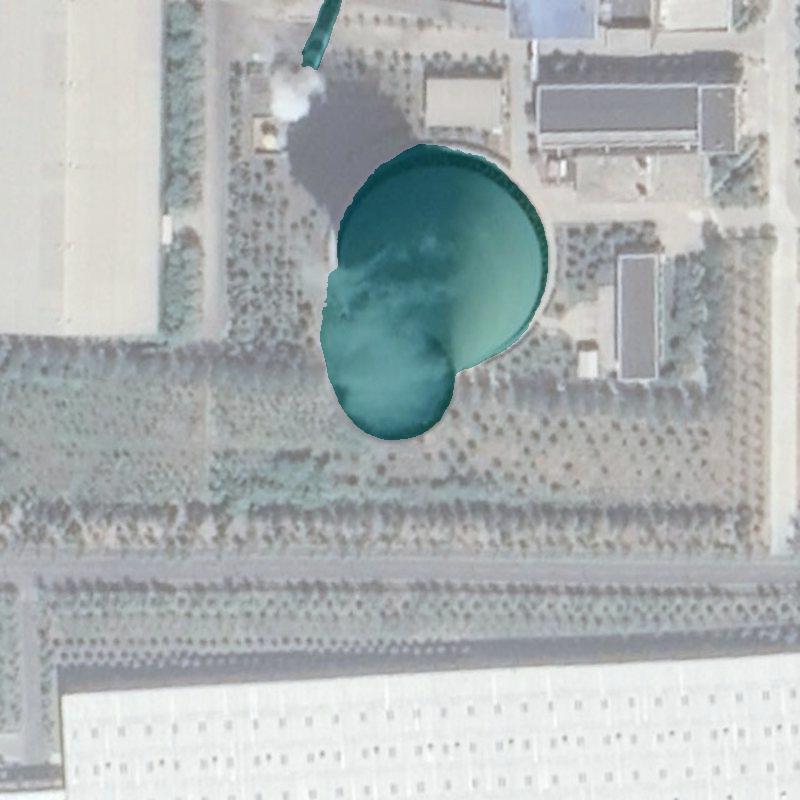}
    \includegraphics[width=0.32\linewidth]{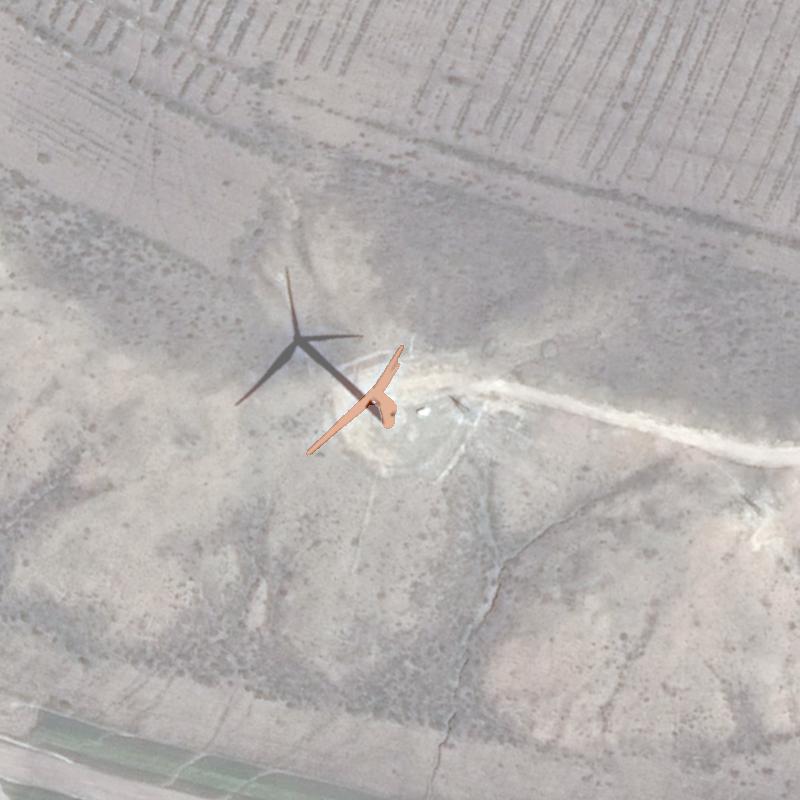}
    \includegraphics[width=0.32\linewidth]{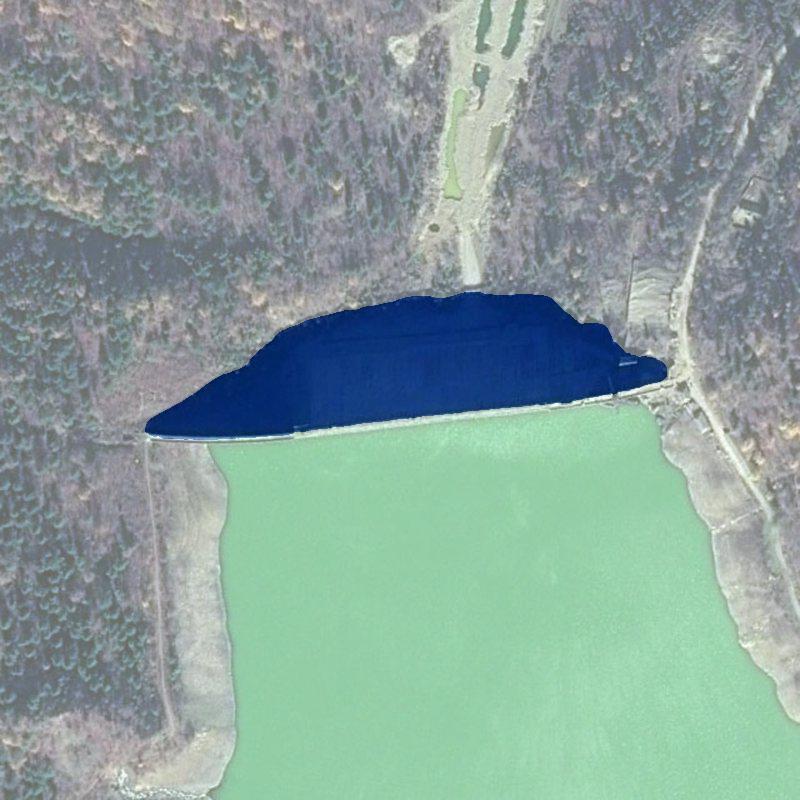}\\
    \includegraphics[width=0.32\linewidth]{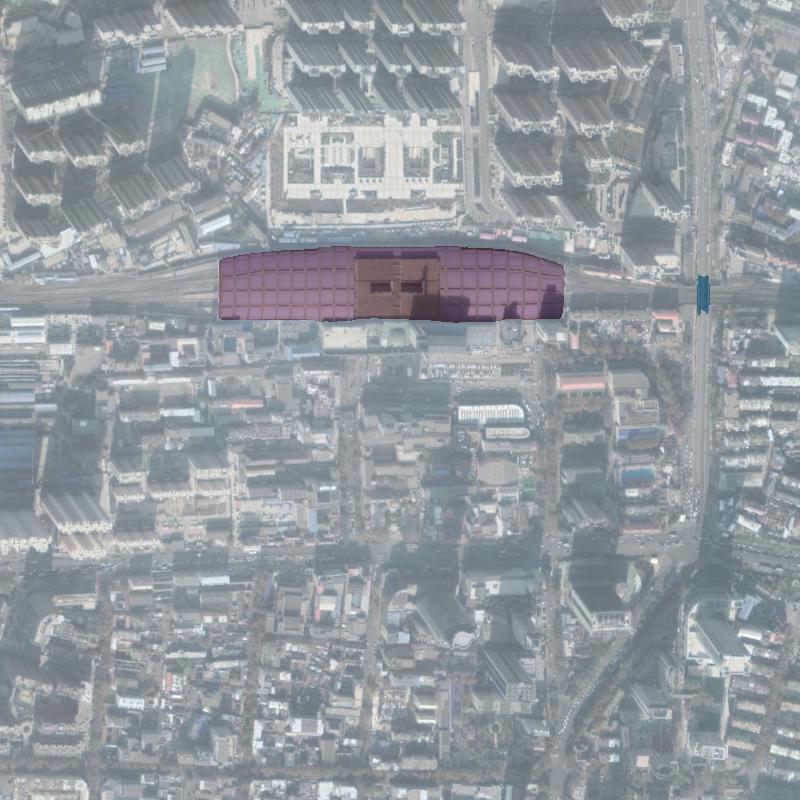}
    \includegraphics[width=0.32\linewidth]{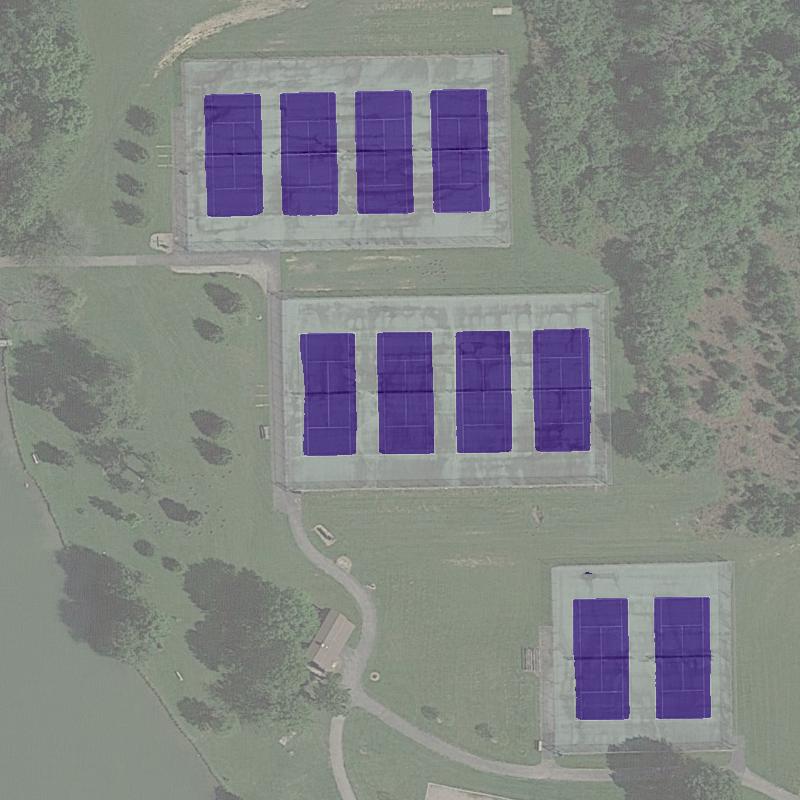}
    \includegraphics[width=0.32\linewidth]{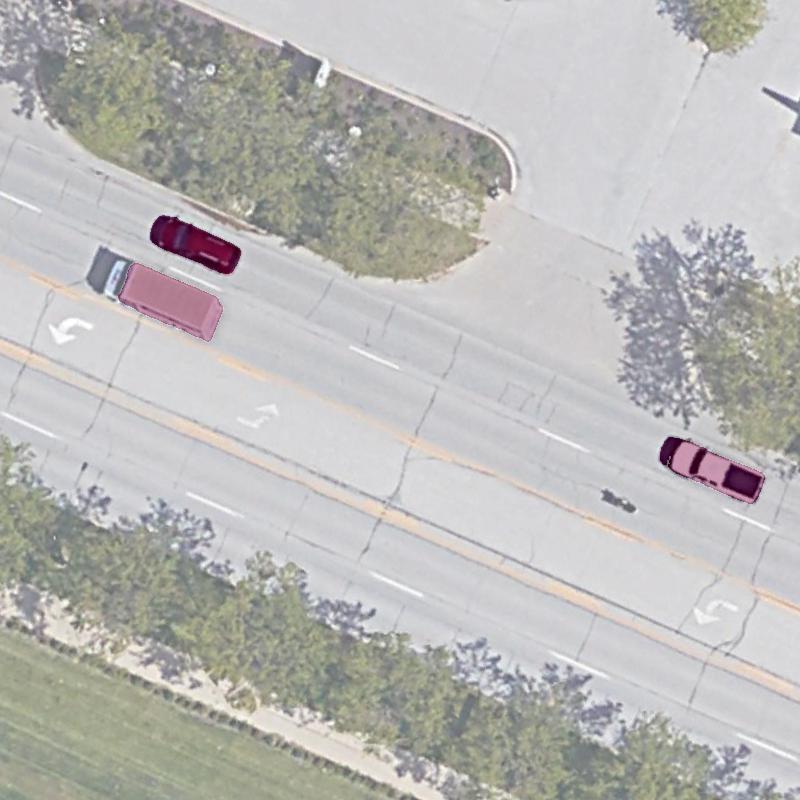}\\
    \caption{Visual examples from the SIOR subset of our SAMRS dataset.}
    \label{vis_sior}
\end{figure}

\begin{figure}[h]
    \centering
    \includegraphics[width=0.32\linewidth]{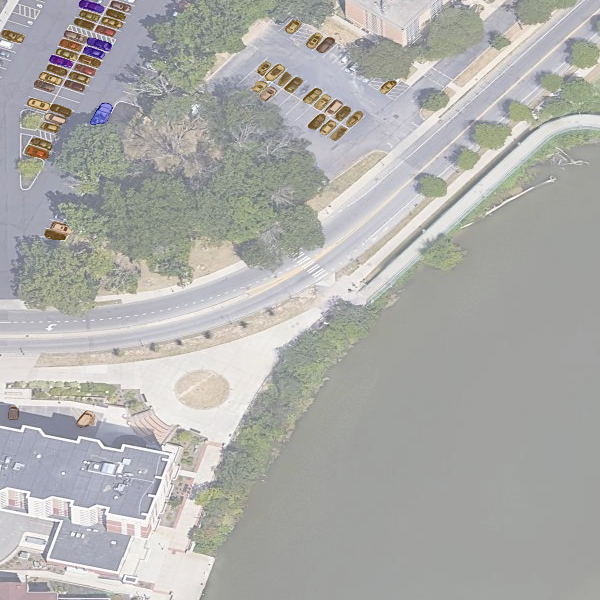}
    \includegraphics[width=0.32\linewidth]{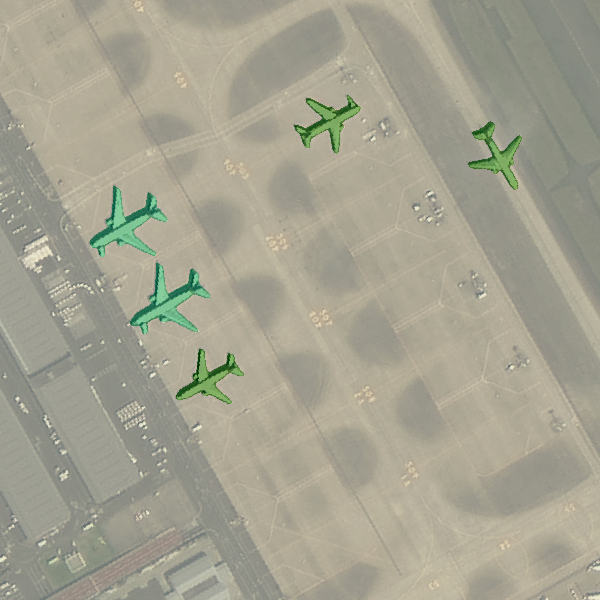}
    \includegraphics[width=0.32\linewidth]{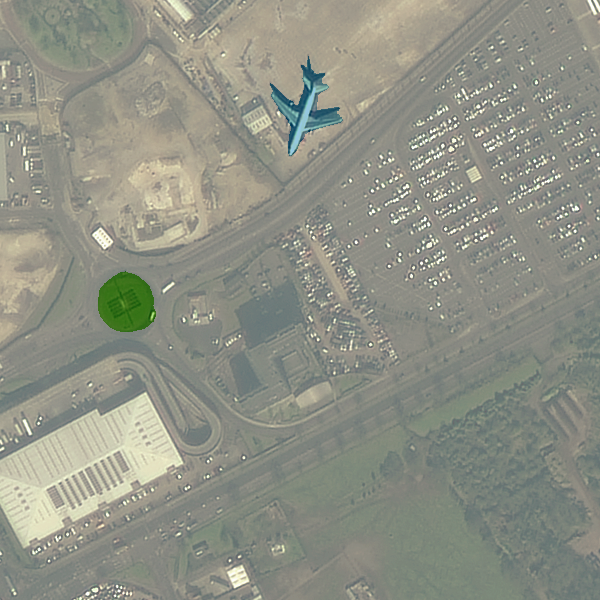}\\
    \includegraphics[width=0.32\linewidth]{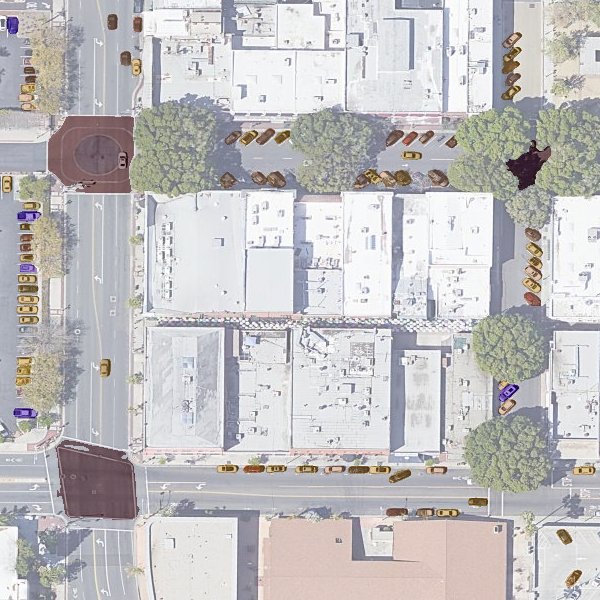}
    \includegraphics[width=0.32\linewidth]{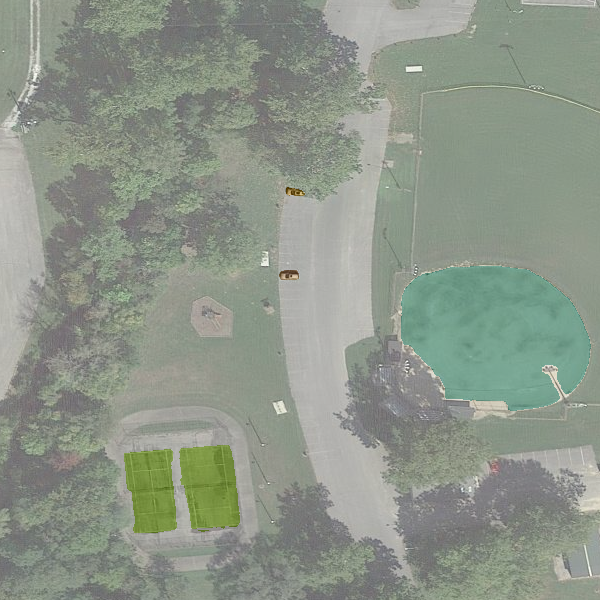}
    \includegraphics[width=0.32\linewidth]{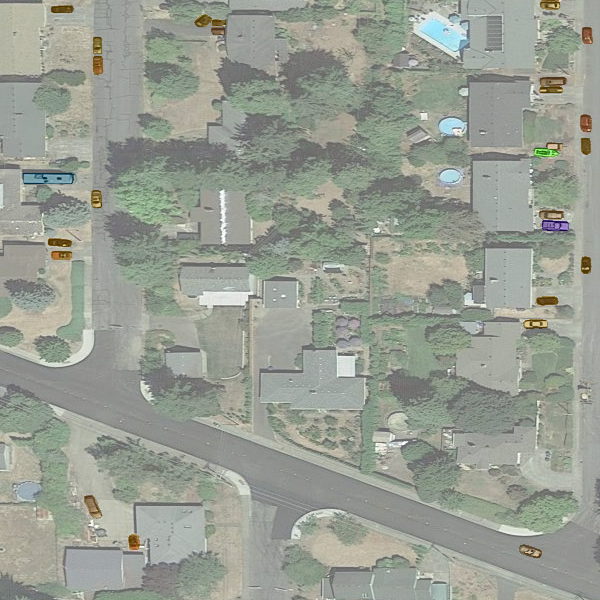}\\
    \includegraphics[width=0.32\linewidth]{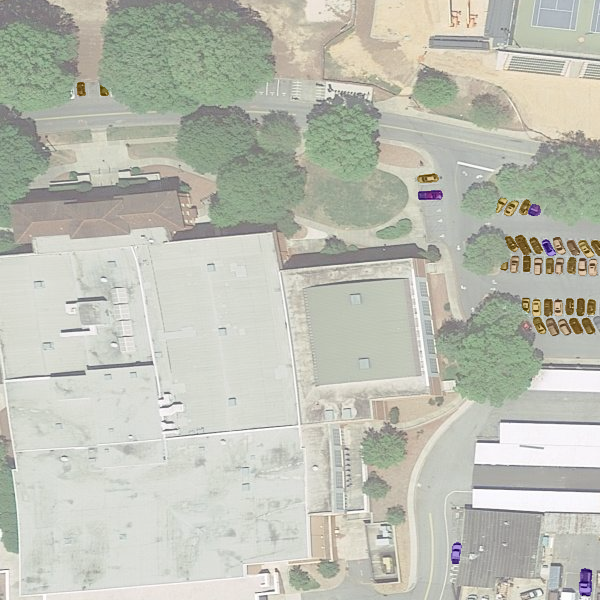}
    \includegraphics[width=0.32\linewidth]{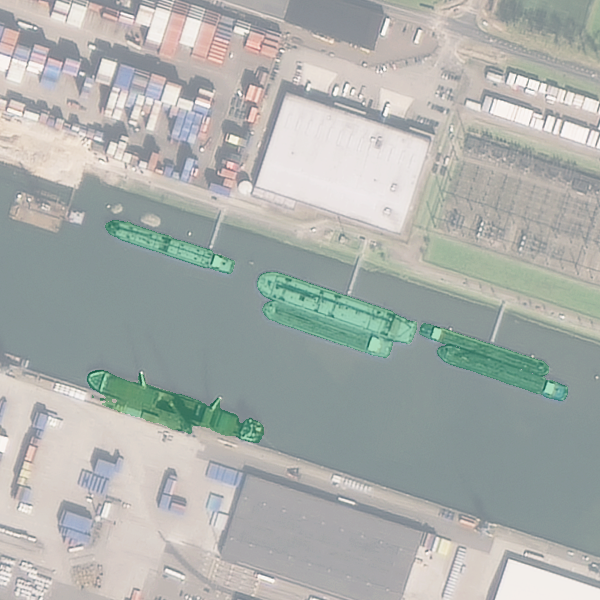}
    \includegraphics[width=0.32\linewidth]{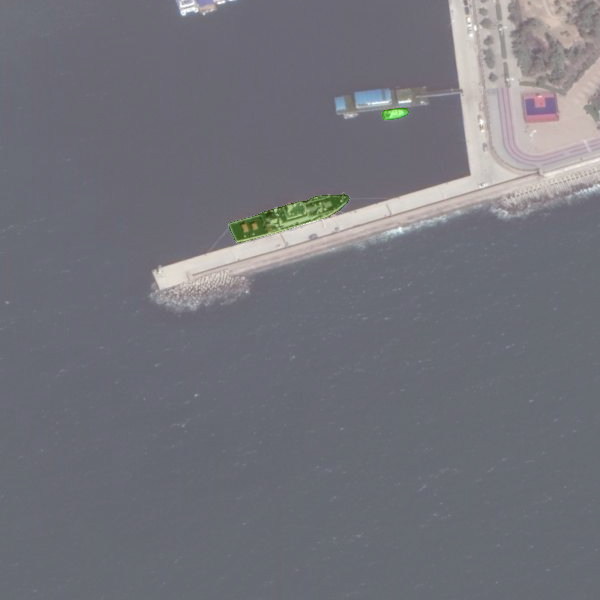}\\
    \includegraphics[width=0.32\linewidth]{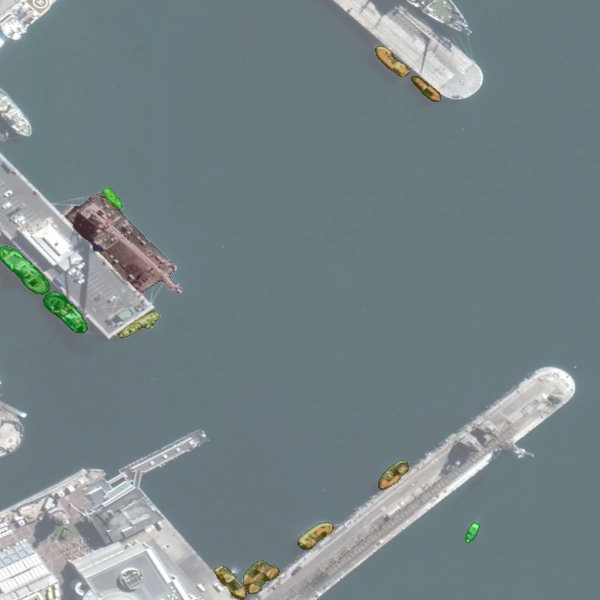}
    \includegraphics[width=0.32\linewidth]{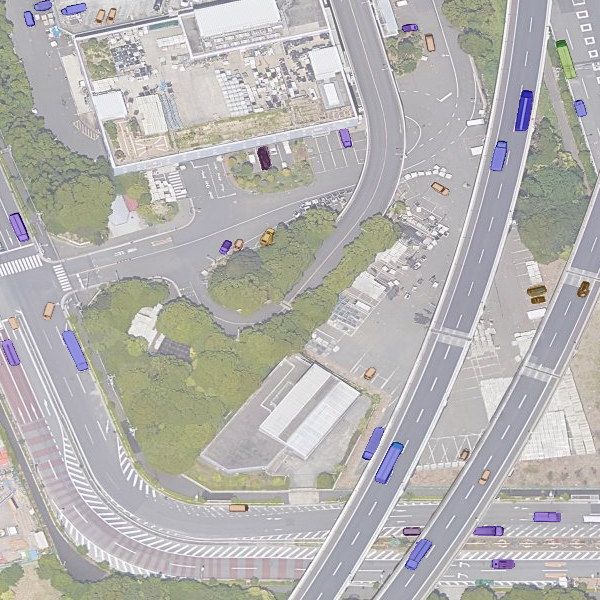}
    \includegraphics[width=0.32\linewidth]{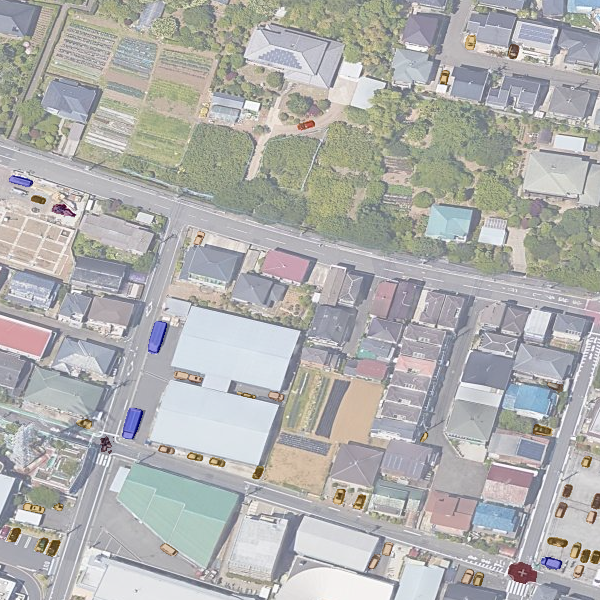}\\
    \caption{Visual examples from the FAST subset of our SAMRS dataset.}
    \label{vis_fast}
\end{figure}

We present more samples of different sets, as shown in Figure \ref{vis_sota}-\ref{vis_fast}

\end{document}